\documentclass[10pt,twocolumn,letterpaper]{article}

\usepackage[pagenumbers]{cvpr}      %

\usepackage{cuted} %

\usepackage{currfile} %

\usepackage{caption} %

\usepackage{microtype}

\makeatletter\renewcommand\paragraph{\@startsection{paragraph}{4}{\z@}{0.4ex}{-0.6em}{\normalfont\normalsize\bfseries}}\makeatother

\usepackage{bm}
\usepackage{algorithm}
\usepackage{algpseudocode}

\usepackage{multirow}
\usepackage{colortbl}
\usepackage{tabularx}
\newcolumntype{Y}{>{\centering\arraybackslash}X}
\newcommand{\tabstyle}{\setlength{\aboverulesep}{0pt}\setlength{\belowrulesep}{0pt}%
  \setlength{\heavyrulewidth}{1.5pt}\setlength{\lightrulewidth}{0.4pt}\renewcommand{\arraystretch}{1.16}}
\newcommand{\oursrow}{\rowcolor{cvprblue!12}}
\usepackage{tikz}
\usepackage{pgfplots}
\pgfplotsset{compat=1.17}
\usepgfplotslibrary{fillbetween}
\usetikzlibrary{arrows.meta,calc,decorations.pathreplacing,shapes.geometric,patterns}
\usepackage{amsthm}

\newtheorem{theorem}{Theorem}
\newtheorem{lemma}{Lemma}

\usepackage{amsmath,amsfonts,bm}

\def\eqref#1{equation~\ref{#1}}

\def\1{\bm{1}}

\def\rmI{{\mathbf{I}}}

\def\vzero{{\bm{0}}}
\def\vone{{\bm{1}}}
\def\vmu{{\bm{\mu}}}

\def\va{{\bm{a}}}

\def\vd{{\bm{d}}}
\def\ve{{\bm{e}}}
\def\vf{{\bm{f}}}

\def\vm{{\bm{m}}}

\def\vo{{\bm{o}}}
\def\vp{{\bm{p}}}

\def\vu{{\bm{u}}}
\def\vv{{\bm{v}}}
\def\vw{{\bm{w}}}

\def\veps{{\bm{\epsilon}}}
\def\vxi{{\bm{\xi}}}
\def\vzeta{{\bm{\zeta}}}

\def\mLambda{{\bm{\Lambda}}}

\DeclareMathAlphabet{\mathsfit}{\encodingdefault}{\sfdefault}{m}{sl}
\SetMathAlphabet{\mathsfit}{bold}{\encodingdefault}{\sfdefault}{bx}{n}

{\theoremstyle{definition}\newtheorem{assumption}{Assumption}}
\newenvironment{restated}[1][]{\par\smallskip\noindent\textbf{#1 (restated).}\ \itshape}{\par\smallskip}

\definecolor{cvprblue}{rgb}{0.21,0.49,0.74}
\usepackage[pagebackref,breaklinks,colorlinks,allcolors=cvprblue]{hyperref}
\usepackage{refcount}
\makeatletter
\newcommand{\savecounter}[1]{\begingroup\protected@edef\@currentlabel{\the\value{#1}}\label{cnt:#1}\endgroup}
\makeatother

\graphicspath{{figures/}}

\def\confName{CVPR}
\def\confYear{2027}

\title{DelayOPSD: On-Policy Self-Distillation from a Synchronous Teacher \\ for Real-time Asynchronous Policies}

\author{
Hyeonjae Kim\textsuperscript{1*}\quad
Jeongsol Kim\textsuperscript{1*}\quad
Sunwoo Park\textsuperscript{1}\quad
Chanyu Moon\textsuperscript{1}\quad
Yoojin Oh\textsuperscript{1}\\
Youngjun Jun\textsuperscript{1}\quad
Kyumin Choi\textsuperscript{2}\quad
Jong Chul Ye\textsuperscript{1\dag}
\\[0.5em]
\textsuperscript{1}Korea Advanced Institute of Science and Technology (KAIST)\qquad
\textsuperscript{2}Sungkyunkwan University\\[0.4em]
{\small\url{https://github.com/present091/DelayOPSD}}
}
\newcommand{\authornotes}{\begingroup\renewcommand\thefootnote{}\footnote{\textsuperscript{*}Equal contribution.\quad \textsuperscript{\dag}Corresponding author.}\addtocounter{footnote}{-1}\endgroup}

\begin{document}
\twocolumn[{\renewcommand\twocolumn[1][]{#1}\maketitle
\begin{center}
\vspace{-2.2em}
\includegraphics[width=\linewidth]{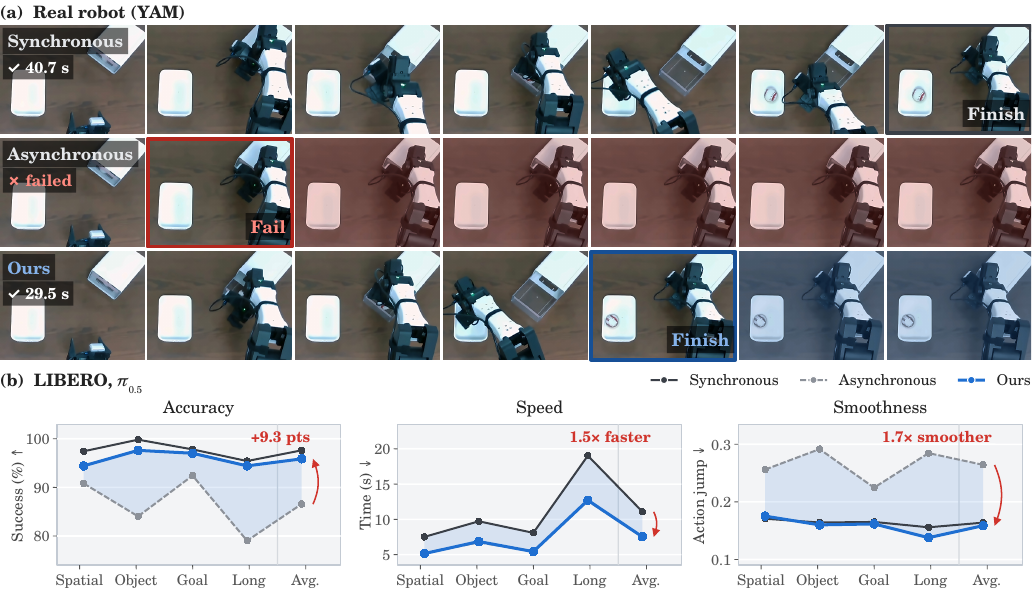}
\vspace{-1.8em}
\captionof{figure}{\textbf{DelayOPSD keeps the speed of asynchronous execution with  success and smoothness close to synchronous execution.} Synchronous execution is accurate and smooth but pauses between action chunks, whereas asynchronous execution removes the pauses but fails more often and jumps at chunk switches.  (a)~Real robot (YAM), 15-step delay, one episode per method. Frames after failure are tinted red and frames after completion blue. Times are averages over successful trials. (b)~LIBERO, $\pi_{0.5}$, 10-step delay. Accuracy is the success rate, Speed is the time of a successful episode including inference, and Smoothness is the action jump at chunk switches.}
\label{fig:teaser}
\vspace{0.1em}
\end{center}}]

\begin{abstract}\authornotes
Flow-based vision-language-action (VLA) policies incur substantial inference latency due to their large vision-language backbones and iterative denoising. 
Asynchronous execution reduces this latency by generating the next action chunk while executing the current one, but the new chunk is conditioned on a stale observation, degrading performance as the delay increases.
We investigate this mismatch by comparing actions predicted from observations at planning and execution times.
We find that these predictions differ little on expert demonstrations but diverge substantially on the policy's own delayed rollouts, particularly in failed episodes.
Motivated by this observation, we propose \textbf{DelayOPSD}, which trains an asynchronous policy to imitate its synchronous counterpart on the policy's own rollouts.
Specifically, a synchronous teacher, obtained from the same policy using the observation at execution time, supervises a delayed student that observes only the earlier frame and the actions still being executed.
The student learns to match the teacher throughout its own denoising process.
Under a 10-step delay, training SmolVLA on demonstrations achieves only 70.2\% success, comparable to naive asynchronous execution (70.4\%), whereas training on its own rollouts improves success to 78.7\%.
Across LIBERO and Kinetix, DelayOPSD achieves the highest average success among the compared methods, without introducing additional modules at deployment.
On a real robot, it further improves success from 46\% to 62\%.

\end{abstract}

\def\frameC{pS!85}\def\frameW{0.7pt}
\definecolor{ovStu}{HTML}{1F4F95}
\definecolor{ovTea}{HTML}{C8384A}
\definecolor{ovPre}{HTML}{5A9BDF}
\definecolor{ovInk}{HTML}{1E1E1E}
\definecolor{cubeC}{HTML}{EB6834}
\colorlet{pS}{ovStu}
\definecolor{vS}{HTML}{153A75}
\definecolor{vT}{HTML}{C8384A}
\definecolor{lossC}{HTML}{C8384A}
\definecolor{infC}{HTML}{5F646B}
\colorlet{cA}{ovPre}
\colorlet{cB}{vS}
\definecolor{wood}{HTML}{EEE4D3}
\newcommand{\arm}[2]{%
  \coordinate (B) at (-0.32,0.00);
  \fill[black!8, opacity=0.9] (-0.32,-0.085) ellipse (0.13 and 0.025);
  \fill[black!70, rounded corners=0.6pt] (-0.41,-0.08) rectangle (-0.23,0.00);
  \coordinate (J1) at #1; \coordinate (J2) at #2; \coordinate (J3) at ($(J2)+(0,-0.06)$);
  \foreach \p/\q in {B/J1, J1/J2, J2/J3} {
    \draw[black!75, line width=4.0pt, line cap=round] (\p) -- (\q);
    \draw[white!96!black, line width=3.1pt, line cap=round] (\p) -- (\q);
    \draw[black!14, line width=0.9pt, line cap=round, transform canvas={yshift=-0.55pt}] (\p) -- (\q);
  }
  \foreach \p in {B, J1, J2} { \fill[black!62] (\p) circle (1.15pt); \fill[black!30] (\p) circle (0.45pt); }
  \fill[black!80, rounded corners=0.4pt] ($(J3)+(-0.06,-0.02)$) rectangle ($(J3)+(0.06,0.01)$);
  \fill[black!80, rounded corners=0.3pt] ($(J3)+(-0.058,-0.02)$) rectangle ($(J3)+(-0.035,-0.085)$);
  \fill[black!80, rounded corners=0.3pt] ($(J3)+(0.035,-0.02)$) rectangle ($(J3)+(0.058,-0.085)$);
}
\def\frameY{0.74}
\newcommand{\frameimg}[3]{%
  \begin{scope}[shift={(#1,\frameY)}]
    \fill[black!12, rounded corners=2.5pt] (-0.50,-0.42) rectangle (0.56,0.34);
    \begin{scope}
      \clip[rounded corners=2pt] (-0.52,-0.38) rectangle (0.52,0.38);
      \shade[top color=white, bottom color=black!7] (-0.52,-0.38) rectangle (0.52,0.38);
      \fill[wood] (-0.52,-0.38) -- (0.52,-0.38) -- (0.52,-0.06) -- (-0.52,-0.06) -- cycle;
      \draw[wood!80!black, line width=0.25pt] (-0.52,-0.06) -- (0.52,-0.06);
      \fill[black!10] (0.11,-0.215) ellipse (0.11 and 0.02); \fill[black!12] (0.37,-0.205) ellipse (0.07 and 0.014);
      \fill[white] (0.10,-0.16) ellipse (0.10 and 0.035);
      \shade[left color=black!15, right color=black!30] (0.00,-0.16) arc (180:360:0.10 and 0.07) -- cycle;
      \shade[top color=cubeC!60, bottom color=cubeC!85!black] (0.30,-0.20) rectangle (0.40,-0.10);
      \fill[cubeC!45] (0.30,-0.10) -- (0.33,-0.07) -- (0.43,-0.07) -- (0.40,-0.10) -- cycle;
      \fill[cubeC!75!black] (0.40,-0.20) -- (0.43,-0.17) -- (0.43,-0.07) -- (0.40,-0.10) -- cycle;
      \begin{scope}[shift={(0,-0.06)}] \arm{#2}{#3} \end{scope}
    \end{scope}
    \draw[\frameC, rounded corners=2pt, line width=\frameW] (-0.52,-0.38) rectangle (0.52,0.38);
  \end{scope}}
\pgfdeclarelayer{background}\pgfsetlayers{background,main}
\begin{figure*}[t]
  \centering
  \includegraphics[width=\textwidth]{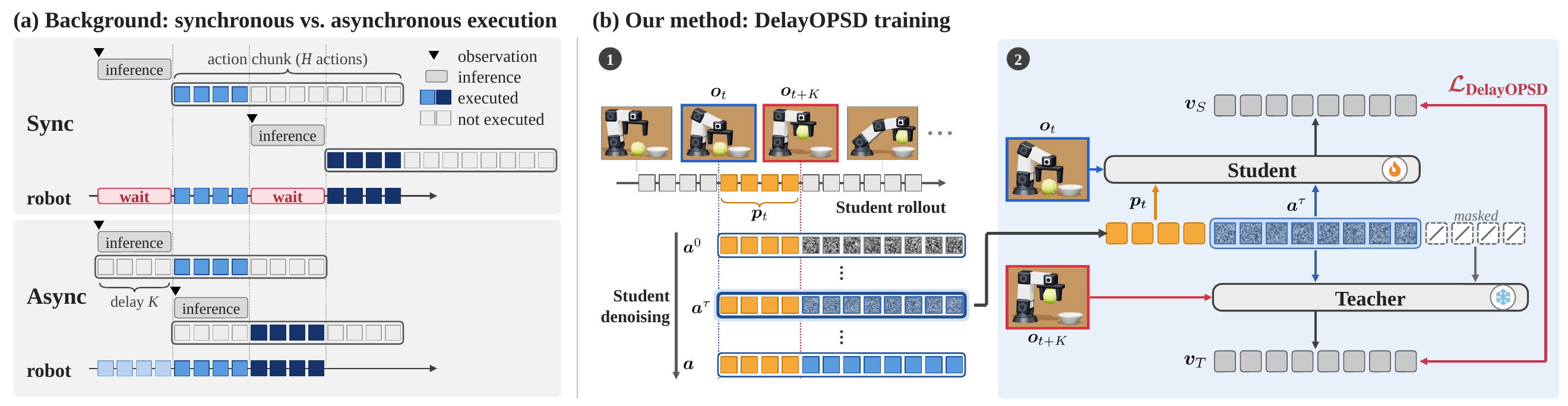}
  \caption{\textbf{Overview of DelayOPSD.} (a) Synchronous execution pauses the robot while each action chunk ($H$ actions) is generated. Asynchronous execution keeps executing the committed actions $\vp_t$ during the $K$ control steps of inference (delay $K$), so the chunk generated from $\vo_t$
  takes effect at $t{+}K$, when the observation is already $\vo_{t+K}$. (b) Training. \textbf{(1) Student states from delayed rollout and denoising:} a step $t$ of
  its own delayed rollout, and an intermediate state $\va^{\tau}$ of its
  own denoising at that step. \textbf{(2) Distillation from the synchronous teacher at these states:} the student
  sees $\vo_t$ and the committed actions $\vp_t$, and the teacher, the same policy
  run without delay, sees $\vo_{t+K}$. Both predict a velocity at
  $\va^{\tau}$, and $\vv_S$ is trained to match $\vv_T$.}
  \label{fig:overview}
  \vspace{-10pt}
\end{figure*}

\section{Introduction}
\label{sec:intro}

Flow-based vision-language-action (VLA) policies such as $\pi_0$~\cite{pi0}, $\pi_{0.5}$~\cite{pi05} and SmolVLA~\cite{smolvla} generate \emph{action chunks}, short sequences of future actions~\cite{act}, by solving a flow ODE initialized with Gaussian noise.
Generating each action chunk requires a forward pass of a large vision-language backbone and multiple denoising steps, taking tens to hundreds of milliseconds on a desktop GPU and even longer on edge or networked hardware~\cite{agouzoul2026async,eaglevla}. 
Under \emph{synchronous execution} the robot stops while each chunk is generated, increasing task completion time and preventing the robot from reacting during generation~\cite{rtc,vlash}. 
\emph{Asynchronous execution}~\cite{rtc} removes these pauses by generating the next chunk while the current chunk is being executed, making it essential for VLA deployment~\cite{rtc,smolvla}.
However, asynchronous execution introduces a mismatch between the observation used for action generation and the state at which the generated actions are executed. While a new chunk is generated, the robot continues executing the \emph{committed actions} of the current chunk for $K$ control steps (the \emph{delay}). Consequently, the new chunk generated from observation $\vo_t$ is executed from step $t+K$, when the environment has already evolved to $\vo_{t+K}$, which was unseen by the policy during generation (\cref{fig:overview}a).
Since flow-based policies trained with behavior cloning do not account for this mismatch, executing them asynchronously leads to a drop in task success rate, which becomes more pronounced as the delay increases (\cref{fig:delay}).

Existing remedies keep the new chunk consistent with the committed actions at inference time~\cite{rtc,bid}, predict the missing observation with an added module~\cite{futurertc}, or fine-tune the policy on demonstrations~\cite{trainingtimertc,legato,softrtc,remac,deflect,rfd}. %
While these approaches mitigate the effects of execution delay, our goal is to directly improve the delayed policy under its own execution dynamics. In particular, the observation-action mismatch can lead the delayed policy to visit states different from those in demonstrations, where the policy may require different actions to recover. In our measurements, the delay barely changes the policy's actions on demonstration states but changes them substantially on its own delayed rollouts, especially in failed episodes (\cref{sec:diagnosis}). This motivates on-policy self-distillation (OPSD), which trains the student on its own rollouts rather than relying solely on demonstration states.

Here, we introduce \textit{DelayOPSD}, an on-policy self-distillation framework for asynchronous VLA execution. Our key insight is to construct an asymmetric teacher-student pair from the \textit{same} policy through different conditioning.
Specifically, the synchronous teacher obtains the execution-time observation $\vo_{t+K}$, while the asynchronous student conditions on $\vo_t$ and the committed actions. By querying the teacher along the student's delayed rollouts, we obtain improved supervision even at states where the student fails, without an external teacher model or ground-truth action annotations. We then distill the teacher into the student by matching their velocity fields along the student's own denoising trajectory, providing dense supervision beyond matching the final actions.

In practice, repeatedly collecting student rollouts online during training is expensive, since it requires interaction with the environment. We therefore collect delayed rollouts once before training and perform distillation offline, as in offline on-policy distillation~\cite{lightningopd,semiopd}. In our experiments, the evaluated online collection variants give no significant gain over this one-time collection. DelayOPSD requires no external expert or additional inference-time module, preserving the original policy architecture and sampler at deployment. Our contributions can be summarized as:

\begin{itemize}
  \item \textbf{Asynchronous execution as on-policy self-distillation from the synchronous policy.} We train the policy on its own delayed rollouts, and its teacher is the same policy run synchronously on the execution-time observation. SmolVLA then reaches 78.7\% at $K{=}10$, against 70.2\% when trained on demonstrations.
  \item \textbf{Distillation on the student's own states in physical and denoising time.} On the student's rollouts, using the student's own samples in denoising time instead of re-noised teacher samples adds 3.4 points of success on average over delays (\cref{sec:ablations}).
  \item \textbf{Highest average success with no module added at deployment.} DelayOPSD has the highest average success of all compared methods on LIBERO (SmolVLA, $\pi_{0.5}$) and Kinetix, without an inference-time module or simulator lookahead at deployment, and raises real-robot success from 46\% to 62\%. It needs no rollouts interleaved with training, which makes training on a real robot practical.
\end{itemize}

\section{Related Work}
\label{sec:related}

\textbf{Asynchronous and real-time chunk execution.}
VLAs~\cite{pi0,pi05,smolvla} denoise chunks of actions~\cite{act}, and
streaming decoding~\cite{flashvla} lowers per-step decoding latency. Asynchronous
methods are compared in a controlled study~\cite{agouzoul2026async}. Inference-time methods leave the policy
unchanged: guided inpainting of the committed actions (real-time chunking, RTC~\cite{rtc}),
initial-noise inversion toward them (PAINT~\cite{paint}), or selection among
sampled chunks by coherence with the previous chunk and contrast with an
early checkpoint (BID~\cite{bid}). Training-time RTC~\cite{trainingtimertc}
fine-tunes the policy to continue a ground-truth prefix. Legato~\cite{legato}
and Soft RTC~\cite{softrtc} replace the hard prefix with soft weights and
blend the previous chunk into every denoising step, and REMAC~\cite{remac}
adds a delay-conditioned LoRA adapter. VLASH~\cite{vlash} fine-tunes on
time-shifted samples to take as input the robot state at
execution time, rolled forward through the committed actions.
FutureRTC~\cite{futurertc} keeps the policy frozen and adds modules that
predict the execution-time latent of its vision encoder and correct the
rolled-forward robot state, and EagleVLA~\cite{eaglevla} predicts the
compressed future VLM (vision-language model) output. A2C2~\cite{a2c2} reads the latest observation
at every step (not compared), and DREAM-Chunk~\cite{dreamchunk} selects among
sampled chunks at each step by matching the observed state to a latent world
model's predictions.

\noindent\textbf{Privileged teachers and interactive imitation.}
DAgger~\cite{dagger} and its delayed-RL variant DIDA~\cite{dida} query an
expert on the states the learner visits. Students without privileged
information are also supervised on their own states, with the waypoints of a
privileged agent~\cite{lbc} or a latent of privileged environment
factors~\cite{rma}. Such a student learns a version of the
teacher's policy with the privileged information marginalized
out~\cite{advisor}. Off-policy imitation with unobserved context can latch
onto past actions~\cite{swamy2022}.

\noindent\textbf{On-policy and self-distillation.}
On-policy distillation evaluates a teacher on student-generated sequences and
minimizes a divergence between their next-token distributions, such as
forward KL, reverse KL or the Jensen--Shannon divergence~\cite{gkd}. On-policy self-distillation
uses one model as student and teacher, with only the teacher given privileged
information~\cite{sdr,pidistill,dopsd}. For diffusion and flow models,
DiffusionOPD~\cite{diffusionopd} matches the student's and teacher's one-step
transition means along the student's own denoising path, and
Flow-OPD~\cite{flowopd} reduces a reverse KL on the student's own samples to
a time-weighted squared error between student and teacher velocities.
Contemporaneous work applies on-policy distillation to VLAs, to consolidate
domain specialists into one policy~\cite{mtopdvla} or to compress a large VLA
into a few-step student~\cite{fastopd}. Both distill from separate teacher
policies. Here the teacher is the same policy given the execution-time
observation, and the student is its delayed version.

\noindent\textbf{Future-conditioned teachers.}
DEFLECT~\cite{deflect} queries a frozen reference policy with the current and
the future observation at demonstration timesteps, and trains the delayed
policy with a flow-matching direct preference optimization (DPO) loss on the two chunks, anchored by flow
matching on demonstrations. Contemporaneous with and independent of our work,
RFD~\cite{rfd} regresses the delayed policy's velocities onto those of
a frozen policy given the later observation, on noised samples of that
policy's chunks at demonstrated transitions, and pairs it with a resolver
that selects among proposed chunks once the later observation arrives. Both train on demonstration states, where the delay barely changes the policy's actions (\cref{sec:diagnosis}). 

In contrast, DelayOPSD trains on the delayed policy's own rollouts, where the delay changes these actions most, and on its own samples in denoising time (\cref{sec:method}).

\section{Preliminaries}
\label{sec:background}

\noindent\textbf{Flow-based VLA policies.}
\label{sec:prelim}
We use superscript $\tau$ to denote the denoising time, where $\tau = 0$ corresponds to noise and $\tau = 1$ to data, while subscript $t$ denotes the control step. For clean actions, we omit the superscript for notational simplicity, unless otherwise stated.

Let $\pi_\theta$ be a policy that, given an observation $\vo_t$ and a language instruction (omitted from the notation below), generates an action chunk $\va_{t:t+H} = [\va_t; \dots; \va_{t+H-1}] \in \mathbb{R}^{H d_a}$ of $H$ actions from Gaussian noise, where $d_a$ denotes the action dimension, $\va_{t}$ is the action for control step $t$, and $\va_{t:t+i}$ the actions for steps $t, \dots, t{+}i{-}1$. The policy is trained by conditional flow matching~\cite{lipman2023flow,liu2023rectified}:
\begin{equation}
  \mathcal{L}_{\mathrm{FM}} = \mathbb{E}\,\big\| \vv_\theta(\va^{\tau}_{t:t+H}, \tau \mid \vo_t) - (\va_{t:t+H} - \veps) \big\|^2,
\end{equation}
where $\va^{\tau}_{t:t+H} = \tau\,\va_{t:t+H} + (1-\tau)\,\veps$ with $\veps \sim \mathcal{N}(\vzero, \rmI)$.
Sampling integrates the flow ODE using $N$ Euler steps,
starting from $\va^0_{t:t+H} \sim \mathcal{N}(\vzero, \rmI)$
and yielding $\va^1_{t:t+H}$ as
\begin{align}
  \va^{\tau+\Delta\tau}_{t:t+H} &= \va^\tau_{t:t+H} + \Delta\tau\, \vv_\theta\big( \va^\tau_{t:t+H}, \tau \mid \vo_t \big),
  \label{eq:euler}
\end{align}
where $\Delta\tau = 1/N$ and
$\tau \in \{0, \Delta\tau, \ldots, 1-\Delta\tau\}$.
Each action in a chunk can also have its own flow time, as in diffusion forcing~\cite{diffusionforcing} and training-time action conditioning~\cite{trainingtimertc}. Actions given as conditioning keep their ground-truth values with $\tau = 1$, and only the remaining actions are denoised from noise. 
Model details and notation are in \cref{sec:impl,tab:notation}.

\noindent\textbf{Asynchronous execution.}
\label{sec:async}
Under asynchronous execution, the robot keeps moving while the policy generates a chunk. The policy replans every $R$ control steps, and generating a chunk takes $K$ control steps, the \emph{delay}. 
We write this schedule as $R{:}K$. At a replanning step $t$ the policy observes $\vo_t$ and starts generating $\va_{t:t+H}$. During the $K$ steps of generation the robot executes the next $K$ actions of the running chunk, the committed actions, which we call the \emph{prefix} $\vp_t \in \mathbb{R}^{K d_a}$ (\cref{fig:overview}a). The new chunk takes effect at step $t{+}K$: its first $K$ actions overlap $\vp_t$ and are discarded, and the rest are executed until the next chunk takes effect.

We call $\vo_t$ the \emph{inference-time observation} and $\vo_{t+K}$,
the observation when the new chunk takes effect, the \emph{execution-time
observation}. $K=0$ is synchronous execution. We consider $K \le R$ (exact schedule in \cref{sec:impl-schedule}). In simulation, time is
frozen during inference, so the delay is exact. On the real robot, $K$ is
fixed above the measured inference latency (\cref{sec:impl-real}). We call the policy fine-tuned for synchronous execution the \emph{base}.
Naive asynchronous execution runs the base under this schedule and
does not take $\vp_t$ as input. We call a policy executed with a delay $K > 0$ the \emph{delayed policy}.

\section{DelayOPSD}
\label{sec:method}

DelayOPSD trains the delayed policy on the states of its own delayed rollouts, failures included, as in on-policy distillation~\cite{gkd}. The delay changes its actions much more at these states than on demonstrations, especially in failed episodes (\cref{tab:gap}). These states have no ground-truth actions, since the actions executed there come from the delayed policy itself and often lead to failure. We therefore use the synchronous policy, given the execution-time observation, as the teacher. It supervises the student on its own states along two time axes (\cref{fig:overview}b). In \emph{physical time}, these are the rollout states at each control step. In \emph{denoising time}, they are the noisy chunks that the student's sampler produces.

\subsection{Distillation from the synchronous teacher}
\label{sec:target}\label{sec:teacher-models}

At a replanning step $t$, the \emph{student} $\pi_\theta(\cdot \mid c_t)$ receives the \emph{context} $c_t = (\vo_t, \vp_t)$, which is the information available at the start of generation,
and $\vp_t$ fills the first $K$ positions of its chunk as conditioning actions with $\tau{=}1$.
The \emph{teacher} $\pi_T$ is the synchronous policy with frozen parameters $\theta_T$. It receives the execution-time observation $\vo_{t+K}$ without a prefix, since the effects of the committed actions are already reflected in $\vo_{t+K}$. In contrast, the student does not observe $\vo_{t+K}$ and is instead trained to leverage $\vp_t$ to infer part of these effects (\cref{sec:prefix}). The student is initialized at the base. The student and the teacher define a distribution over the same actions, those for steps $t{+}K$ to $t{+}H{-}1$:
\begin{equation}
\begin{aligned}
  &\pi_\theta(\va_{t+K:t+H} \mid c_t) && \text{(student)}, \\
  &\pi_T(\va_{t+K:t+H} \mid \vo_{t+K}) && \text{(teacher)}.
\end{aligned}
  \label{eq:pair}
\end{equation}
The teacher's chunk starts at $t{+}K$, so only its first $H{-}K$ actions overlap the student's chunk, and the inputs at its last $K$ positions are masked. For SmolVLA, whose action tokens attend causally, this masking does not change the first $H{-}K$ outputs, and the frozen base is the
teacher ($\theta_T = \theta_{\mathrm{base}}$). For $\pi_{0.5}$ and the Kinetix policy, whose
action tokens mix non-causally, the base is first fine-tuned briefly with the
same masking on its training demonstrations and then frozen (\cref{sec:impl}). Its synchronous success is unchanged (97.7\% vs.\ 97.6\% for $\pi_{0.5}$).

\label{sec:pathloss}DelayOPSD matches the two velocity fields at the states of the student's own sampler, which determine its actions at deployment (\cref{sec:denoising}). Let $\va^{\tau}$ denote the student's noisy actions $\va^{\tau}_{t+K:t+H}$ after the prefix. From $\va^{0} = \veps$, the student's Euler steps (\cref{eq:euler}) give the states $\va^{\tau}$ at $\tau \in \{0, \Delta\tau, \ldots, 1-\Delta\tau\}$, computed without gradient. At these states, the two velocities are
\begin{equation}
\begin{aligned}
  \vv_S(\va^{\tau}, \tau \mid c_t) &= \vv_\theta\big([\vp_t; \va^{\tau}], \tau \mid \vo_t\big)_{t+K:t+H} \\
  \vv_T(\va^{\tau}, \tau \mid \vo_{t+K}) &= \vv_{\theta_T}\big([\va^{\tau}; \vm], \tau \mid \vo_{t+K}\big)_{t+K:t+H}
\end{aligned}
  \label{eq:velocities}
\end{equation}
where $\vm$ denotes the $K$ masked inputs and the subscript selects the outputs for these control steps (details in \cref{sec:impl-schedule}). The loss is
\begin{equation}
  \mathcal{L}_{\mathrm{DelayOPSD}} = \mathbb{E}_{(K, c_t, \vo_{t+K}) \sim \mathcal{B},\, \veps,\, \tau} \big\| \vv_S - \mathrm{sg}(\vv_T) \big\|_{\vw}^2,
  \label{eq:opsd}
\end{equation}
where $\mathcal{B} = \bigcup_K \mathcal{B}_K$ are the buffers of delayed rollouts (\cref{sec:physical}, sampling in \cref{sec:loss-norm}), $\veps \sim \mathcal{N}(\vzero, \rmI)$, $\tau \sim \mathcal{U}\{0, \Delta\tau, \ldots, 1-\Delta\tau\}$, both velocities are evaluated at $(\va^{\tau}, \tau)$, and $\mathrm{sg}$ is the stop-gradient. The norm $\|\cdot\|_{\vw}$ down-weights later control steps to emphasize near-future actions, with a normalization shared by all delays (\cref{sec:loss-norm}). \Cref{alg:strap} summarizes training.

At deployment\label{sec:deploy} the student runs alone under $R{:}K$, without guidance~\cite{rtc} or an added module.

\definecolor{algNote}{HTML}{7A7A75}
\algrenewcommand\algorithmicrequire{\textbf{Input:}}
\algrenewcommand\algorithmicensure{\textbf{Output:}}
\algrenewcommand\algorithmiccomment[1]{\hfill\textcolor{algNote}{$\triangleright$ #1}}
\begin{algorithm}[t]
  \caption{DelayOPSD}
  \label{alg:strap}
  \small
  \begin{algorithmic}[1]
    \Require base $\theta_{\mathrm{base}}$, frozen teacher $\theta_T$, delays $\mathcal{K}$, learning rate $\eta$
    \Ensure student $\theta$
    \For{$K \in \mathcal{K}$} \Comment{collect once, \cref{sec:physical}}
      \State Run $\theta_{\mathrm{base}}$ under naive $R{:}K$ and store $(\vo_t, \vp_t, \vo_{t+K})$ in $\mathcal{B}_K$
      \State Optional: drop failures in scenes the base fails at $K{=}0$
    \EndFor
    \State $\theta \gets \theta_{\mathrm{base}}$ \Comment{train}
    \Repeat
      \State Sample $K \in \mathcal{K}$ and $(\vo_t, \vp_t, \vo_{t+K}) \sim \mathcal{B}_K$
      \State Sample $\tau \sim \mathcal{U}\{0, \Delta\tau, \ldots, 1-\Delta\tau\}$
      \State $\va^{\tau} \gets$ Euler state of $\pi_\theta(\cdot \mid c_t)$ at $\tau$ \Comment{no grad.}
      \State $\vv_S \gets \vv_S(\va^{\tau}, \tau \mid c_t)$ \Comment{student}
      \State $\vv_T \gets \vv_T(\va^{\tau}, \tau \mid \vo_{t+K})$ \Comment{teacher}
      \State $\mathcal{L} \gets \| \vv_S - \mathrm{sg}(\vv_T) \|_{\vw}^2$ \Comment{\cref{eq:opsd}}
      \State $\theta \gets \theta - \eta \nabla_\theta \mathcal{L}$
    \Until{converged}
    \State \Return $\theta$ \Comment{deployed as is}
  \end{algorithmic}
\end{algorithm}

\subsection{Training states from delayed rollouts}
\label{sec:physical}\label{sec:train}

For each delay $K$, DelayOPSD runs the base under naive $R{:}K$ execution and
records at every replanning step $t$ a tuple $(\vo_t, \vp_t, \vo_{t+K})$: the
student's inputs and the teacher's input. The executed actions are
never used as targets.

\paragraph{Collected once.} As in on-policy distillation, the teacher is queried on the states the delayed policy itself reaches, not on the demonstrator's. Collecting these states anew during training would require new VLA rollouts between updates. This is costly in simulation and impractical on a robot, where every rollout is a physical episode. DelayOPSD therefore collects a fixed buffer $\mathcal{B}_K$
per delay once, before training, by running the base, which is the student at initialization.
In language models, distilling on rollouts of the initial student collected once is known as offline on-policy distillation~\cite{lightningopd} or
semi-on-policy distillation~\cite{semiopd}, and often matches or
outperforms online collection~\cite{semiopd}. Collecting the buffer online gives no significant gain (\cref{tab:ablations}d).

\paragraph{Failed episodes.} The buffer keeps the delayed policy's failed episodes. Failed episodes often end with the robot standing still until timeout, and we keep only the first 30 steps after the robot stops (\cref{sec:impl}). For some failed episodes, the base fails even when run synchronously from the same initial scene. The teacher's targets are less accurate in such scenes, so on LIBERO a \emph{scene filter} removes such episodes from the buffer. In the remaining scenes, the teacher recovers from failed states more often than the delayed policy (\cref{sec:ablations}). The filter removes 53.8\% of failed episodes for SmolVLA, whose synchronous success is 83.4\%, and 22.2\% for $\pi_{0.5}$, whose synchronous success is 97.6\%. At $K{=}10$, the filter raises success by 2.5 points for SmolVLA and 0.1 points for $\pi_{0.5}$.

\subsection{Theoretical analysis}
\label{sec:why-onpolicy}

We analyze the student-teacher gap with a trajectory-level reverse KL, in physical and then in denoising time.

\paragraph{Physical time (replanning steps).} Running the teacher at every replanning step on the execution-time observation defines a reference policy $\pi^*$ with the same
$R{:}K$ timing (\cref{sec:process}). Let $P_\theta$ and $P_*$ be the
trajectory distributions of the student and of $\pi^*$ over the $M$
replanning steps of an episode. By Pinsker's inequality, the difference in success between the two is at most a constant times the square root of their KL, in either direction. The delay changes the actions most on the delayed policy's own trajectories  (\cref{tab:gap}). We therefore use the reverse direction, $\mathrm{KL}(P_\theta \| P_*)$, which is an expectation under these trajectories and can be estimated from the student's rollouts.

\begin{theorem}
\label{thm:rollout}
For the process of \cref{sec:process}, if the student's sampler noise is
independent of the environment and both chunk distributions have positive
densities, then with $\pi_\theta(\cdot \mid c_t)$ and
$\pi_T(\cdot \mid \vo_{t+K})$ the distributions of \cref{eq:pair},
\begin{equation}
  \mathrm{KL}(P_\theta \,\|\, P_*) = \sum_{t}
  \mathbb{E}_{P_\theta}\Big[\mathrm{KL}\big(\pi_\theta(\cdot \mid c_t)
  \,\big\|\, \pi_T(\cdot \mid \vo_{t+K})\big)\Big],
  \label{eq:rollout}
\end{equation}
where the sum runs over the $M$ replanning steps $t$.
\end{theorem}

In \cref{eq:rollout}, each term is averaged over the contexts that the
student reaches, not over those the demonstrator reaches. \Cref{tab:ablations}a compares training on these contexts with training on demonstrations. \Cref{thm:rollout} requires only that the student generates its chunk from $c_t$, without access to $\vo_{t+K}$
(\cref{sec:rollout-proof}).

\paragraph{Denoising time (sampler steps).}\label{sec:denoising} The chunk-level terms are KL divergences between flow policies, whose
densities are costly to evaluate. \Cref{thm:euler} bounds each of them by a
velocity error and specifies where this error is evaluated. The velocities and the norm are those of \cref{eq:velocities,eq:opsd}.

\begin{theorem}
\label{thm:euler}
Let $\va^{\tau}$ be the student's denoising path ($N$ Euler steps from noise) and
$\ell(c_t, \vo_{t+K}) = \mathbb{E}_{\veps, \tau}\|\vv_S(\va^{\tau}, \tau \mid c_t) - \vv_T(\va^{\tau},
\tau \mid \vo_{t+K})\|_{\vw}^2$, $\tau \sim \mathcal{U}\{0, \Delta\tau, \ldots, 1-\Delta\tau\}$. Under the
smoothness conditions of \cref{sec:euler-kl-proof}, with Lipschitz ratio
$\rho < 1$ and score-gap constant $B$,
\begin{equation}
  \mathrm{KL}\big(\pi_\theta(\cdot \mid c_t) \,\big\|\, \pi_T(\cdot \mid \vo_{t+K})\big)
  \le \sqrt{B\, \ell(c_t, \vo_{t+K}) / (1 - \rho)} .
  \label{eq:euler-kl}
\end{equation}
\end{theorem}

The bound uses the velocity error at the states of the student's own sampler, as \cref{eq:opsd} does. \Cref{tab:ablations}a compares matching at these states with
re-noised teacher samples.

\paragraph{From the bounds to training.} \Cref{thm:euler} and Jensen's inequality bound the trajectory KL of \cref{thm:rollout} by an increasing
function of the velocity loss averaged over the student's contexts
(\cref{sec:return-proof}). The bound holds under the assumptions of \cref{sec:euler-kl-proof}. In training, this loss is evaluated at the contexts of the delayed-rollout buffer (\cref{sec:physical}). This buffer holds rollouts of the base, which is the student at initialization, and re-collecting it during training gives no significant gain (\cref{tab:ablations}d). Each update holds the buffer states fixed, so the loss is a surrogate for this KL (\cref{sec:surrogate}).

\paragraph{The role of the committed actions.}\label{sec:prefix}

The committed actions $\vp_t$ determine the robot's motion during the delay, so they partly reveal how the scene changes by $t{+}K$.
The student still does not see $\vo_{t+K}$, so the loss is minimized by $\bar\vv_T$,
the teacher's velocity averaged over the observations $\vo_{t+K}$ consistent with
$c_t$.
We call the infimum of each term of \cref{eq:rollout} over all students that see only $c_t$ the \emph{information floor}. The floor is zero when
$c_t$ determines $\vo_{t+K}$, and conditioning on $\vp_t$ can only lower it
(\cref{sec:floor}, ablation in \cref{tab:ablations}c).

\section{Experiments}
\label{sec:experiments}

\begin{table*}[!t]
  \centering
  \footnotesize
  \caption{\textbf{One teacher, different training states.}
  LIBERO, four-suite average success (\%) at $K{=}10$, with DelayOPSD in bold. All else is shared:
  teacher, loss, updates. (a) Columns: physical states from demonstrations, synchronous rollouts or delayed rollouts. Rows: denoising states. $K{=}5$, RFD's training states~\cite{rfd} and more $\pi_{0.5}$ rows are in \cref{tab:crossing,tab:ablations-k5}. Filtered: successes, and failures in
  scenes the base solves under synchronous execution. $^*$First-round buffer plus rollouts of the trained
  student, filtered. $^\S$Collected during training, filtered.}
  \label{tab:ablations}
  \setlength{\tabcolsep}{3pt}
  \begin{tabular}[t]{@{}lccc@{}}
    \multicolumn{4}{@{}c@{}}{\textbf{(a) Physical $\times$ denoising states, SmolVLA}}\\[2pt]
    \toprule
    Denoising states & Demos & Sync. & \textbf{Delayed} \\
    \midrule
    Teacher, re-noised & 70.90 & 73.05 & 73.50 \\
    Student, re-noised & 70.05 & 74.65 & 76.95 \\
    \textbf{Student path} & 70.15 & 75.45 & \textbf{78.70} \\
    \bottomrule
  \end{tabular}\hfill
  \begin{tabular}[t]{@{}lcc@{}}
    \multicolumn{3}{@{}c@{}}{\textbf{(b) Episodes kept}}\\[2pt]
    \toprule
    Episodes & SmolVLA & $\pi_{0.5}$ \\
    \midrule
    Successes & 74.35 & 95.45 \\
    Failures & 66.30 & 94.05 \\
    All & 76.25 & 95.75 \\
    \textbf{Filtered} & \textbf{78.70} & \textbf{95.85} \\
    \bottomrule
  \end{tabular}\hfill
  \begin{tabular}[t]{@{}lcc@{}}
    \multicolumn{3}{@{}c@{}}{\textbf{(c) Student input}}\\[2pt]
    \toprule
    Input & SmolVLA & $\pi_{0.5}$ \\
    \midrule
    $\vo_t$ & 76.10 & 91.65 \\
    \textbf{$\vo_t$ + prefix} & \textbf{78.70} & \textbf{95.85} \\
    \bottomrule
  \end{tabular}\hfill
  \begin{tabular}[t]{@{}lcc@{}}
    \multicolumn{3}{@{}c@{}}{\textbf{(d) Collection}}\\[2pt]
    \toprule
    Buffer & SmolVLA & $\pi_{0.5}$ \\
    \midrule
    \textbf{Once} & \textbf{78.70} & \textbf{95.85} \\
    Two rounds$^*$ & 76.15 & 95.80 \\
    Online$^\S$ & 76.45 & 95.95 \\
    \bottomrule
  \end{tabular}
  \vspace{-6pt}
\end{table*}

\begin{table*}[t]
  \centering
  \small
  \setlength{\tabcolsep}{4.5pt}
  \tabstyle
  \caption{\textbf{LIBERO, SmolVLA, delay sweep.} Average success rate (\%)
  over the four suites under $10{:}K$ execution. Synchronous execution ($K{=}0$), the reference for all
  delays, reaches 83.35\%. Avg.: $K{=}1$--10. Bold: best, underline: second best. Baseline settings: \cref{tab:impl}. $^\dagger$Reads the execution-time robot state
  from the simulator. Its fine-tuning lowers synchronous success to 77.2\%. Per suite: \cref{sec:suites}.}
  \label{tab:smolvla-k}
  \begin{tabularx}{\textwidth}{l*{10}{Y}|Y}
    \toprule
    Method & $K{=}1$ & 2 & 3 & 4 & 5 & 6 & 7 & 8 & 9 & 10 & Avg. \\
    \midrule
    SmolVLA, naive async & 82.10 & 80.70 & 81.05 & 78.90 & 77.70 & 77.00 & 75.10 & 73.95 & 72.85 & 70.40 & 76.98 \\
    Inference-time RTC~\cite{rtc} & 81.70 & \underline{81.95} & \underline{82.25} & \underline{79.55} & 77.40 & 76.95 & 75.55 & 73.95 & 72.80 & 70.90 & 77.30 \\
    Training-time RTC~\cite{trainingtimertc} & 80.90 & 78.75 & 78.70 & 78.30 & 77.25 & \underline{77.30} & 77.00 & 75.10 & 74.90 & 75.05 & 77.33 \\
    VLASH~\cite{vlash}$^\dagger$ & 78.40 & 76.10 & 76.35 & 75.55 & 75.55 & 74.50 & 73.35 & 72.95 & 71.60 & 70.80 & 74.52 \\
    DEFLECT~\cite{deflect} & 81.70 & 81.70 & 81.30 & 79.40 & \underline{79.05} & 76.40 & 75.80 & 72.55 & 73.15 & 70.55 & 77.16 \\
    FutureRTC~\cite{futurertc} & \underline{82.80} & 81.60 & 79.75 & 79.25 & 78.55 & 77.10 & \underline{77.55} & \underline{76.60} & \underline{76.55} & \underline{75.50} & \underline{78.53} \\
    \midrule
    \oursrow DelayOPSD (ours) & \textbf{83.15} & \textbf{82.05} & \textbf{82.30} & \textbf{82.35} & \textbf{81.55} & \textbf{80.05} & \textbf{79.90} & \textbf{79.40} & \textbf{77.70} & \textbf{78.70} & \textbf{80.72} \\
    \bottomrule
  \end{tabularx}
  \vspace{-10pt}
\end{table*}

\begin{table*}[t]
  \centering
  \small
  \setlength{\tabcolsep}{4.5pt}
  \tabstyle
  \caption{\textbf{LIBERO, $\pi_{0.5}$, delay sweep}. Average success
  rate (\%) over the four suites under $10{:}K$ execution, 500 episodes per
  suite. Synchronous base: 97.60\% (teacher: 97.70\%). Bold: best, underline: second best. Baseline settings: \cref{sec:impl-pi05}. $^\dagger$Reads the execution-time robot state
  from the simulator (synchronous: 87.15\%).
  Per suite: \cref{sec:suites}.}
  \label{tab:libero-main}
  \begin{tabularx}{\textwidth}{l*{10}{Y}|Y}
    \toprule
    Method & $K{=}1$ & 2 & 3 & 4 & 5 & 6 & 7 & 8 & 9 & 10 & Avg. \\
    \midrule
    $\pi_{0.5}$, naive async & 97.55 & 97.20 & 96.35 & 95.70 & 94.30 & 92.45 & 92.05 & 90.90 & 88.40 & 86.55 & 93.15 \\
    Inference-time RTC~\cite{rtc} & 97.60 & \underline{97.75} & 95.65 & 95.65 & 94.40 & 94.00 & 92.40 & 90.00 & 88.40 & 87.65 & 93.35 \\
    Training-time RTC~\cite{trainingtimertc} & 97.05 & 97.70 & 96.75 & 96.65 & 95.70 & 94.45 & 94.00 & 93.25 & 92.55 & 90.55 & 94.87 \\
    VLASH~\cite{vlash}$^\dagger$ & 87.95 & 89.70 & 90.00 & 89.65 & 88.35 & 86.80 & 85.40 & 83.80 & 80.85 & 76.35 & 85.89 \\
    DEFLECT~\cite{deflect} & \textbf{97.75} & \textbf{97.90} & 97.05 & 96.70 & 95.15 & 93.95 & 92.90 & 90.75 & 89.95 & 88.60 & 94.07 \\
    FutureRTC~\cite{futurertc} & 97.45 & 97.65 & \textbf{97.50} & \underline{97.05} & \underline{96.65} & \underline{96.50} & \underline{96.30} & \underline{95.55} & \underline{95.20} & \underline{95.25} & \underline{96.51} \\
    \midrule
    \oursrow DelayOPSD (ours) & \underline{97.65} & 97.30 & \underline{97.10} & \textbf{97.85} & \textbf{97.15} & \textbf{97.35} & \textbf{96.75} & \textbf{97.20} & \textbf{96.55} & \textbf{95.85} & \textbf{97.08} \\
    \bottomrule
  \end{tabularx}
  \vspace{-4pt}
\end{table*}

\begin{table*}[t]
  \centering
  \small
  \tabstyle
  \caption{\textbf{Kinetix.} Mean solved rate (\%) over 12 levels by delay
  $K$ (95\% CI of the mean $\le \pm 0.6$). Avg.\ over $K{=}0$--$4$. Bold: best, underline: second best. $^\dagger$Future robot state from a
  simulator rollout of the committed actions at deployment (VLASH, DEFLECT on
  the VLASH checkpoint, and FutureRTC, which predicts only the environment part). Hard: RTC without soft weights. All rows are our evaluations with $R = \max(K, 1)$. $^a$Our implementation, uniform
  chunk averaging. $^b$32 epochs. $^c$Retrained with the released code (no released checkpoint). $^d$Released weights.
  $^e$Our implementation of the paper (no released code) on the released VLASH checkpoint.}
  \label{tab:kinetix}
  \begin{tabular}{ll*{5}{c}|c}
    \toprule
    Group & Method & $K{=}0$ & $K{=}1$ & $K{=}2$ & $K{=}3$ & $K{=}4$ & Avg. \\
    \midrule
    \multirow{5}{*}{Inference-time}
      & Naive async & 91.5 & 76.4 & 70.7 & 58.2 & 49.0 & 69.2 \\
      & TE~\cite{act}$^a$ & 74.2 & 63.7 & 57.4 & 52.9 & 52.1 & 60.1 \\
      & BID~\cite{bid} & \underline{91.6} & 87.6 & 81.2 & 68.4 & 55.5 & 76.9 \\
      & RTC (hard)~\cite{rtc} & 91.4 & 84.6 & 81.4 & 73.1 & 61.2 & 78.3 \\
      & RTC~\cite{rtc} & \textbf{92.1} & 89.3 & 84.4 & 74.7 & 61.1 & 80.3 \\
    \midrule
    \multirow{2}{*}{Training-time}
      & T-RTC~\cite{trainingtimertc}$^b$ & 89.0 & 88.6 & 86.4 & 81.8 & 75.5 & 84.3 \\
      & REMAC~\cite{remac}$^c$ & 89.9 & 89.0 & 86.2 & 80.3 & 73.9 & 83.9 \\
    \midrule
    \multirow{3}{*}{Privileged$^\dagger$}
      & VLASH~\cite{vlash}$^{\dagger d}$ & 90.1 & 89.1 & 87.6 & 85.7 & 81.5 & 86.8 \\
      & DEFLECT~\cite{deflect}$^{\dagger e}$ & 90.7 & 89.9 & 88.7 & \underline{87.7} & 84.6 & 88.3 \\
      & FutureRTC~\cite{futurertc}$^{\dagger d}$ & 91.4 & \underline{90.6} & \underline{89.5} & 87.4 & \underline{85.9} & \underline{89.0} \\
    \midrule
    \oursrow Ours & DelayOPSD & 91.2 & \textbf{90.8} & \textbf{89.9} & \textbf{88.7} & \textbf{86.5} & \textbf{89.4} \\
    \bottomrule
  \end{tabular}
  \vspace{-10pt}
\end{table*}

\definecolor{cStrap}{HTML}{2A78D6}
\definecolor{cTrain}{HTML}{EB6834}
\definecolor{cPriv}{HTML}{8E5BC4}
\pgfplotsset{
  delayplot/.style={
    scale only axis, width=0.37\linewidth, height=2.8cm,
    axis lines=left, axis line style={-, black!45},
    tick style={black!45}, tick label style={font=\scriptsize, black!70},
    xlabel style={font=\scriptsize, black!70, yshift=3pt},
    ylabel style={font=\scriptsize, black!70, yshift=-4pt},
    title style={font=\footnotesize\bfseries, yshift=-6pt},
    ymajorgrids, grid style={black!10, line width=0.3pt},
    clip=false, enlarge x limits=0.03,
  },
  m/.style={line width=0.7pt, mark size=1.3pt, mark options={solid}},
  naive/.style={m, black!45, mark=o},
  te/.style={m, black!30, mark=triangle},
  bid/.style={m, black!38, mark=diamond},
  rtch/.style={m, black!55, mark=x},
  rtc/.style={m, black!70, mark=square},
  trtc/.style={m, cTrain, mark=square*},
  remac/.style={m, cTrain!85!black, mark=diamond*},
  vlash/.style={m, cPriv!55, densely dashed, mark=o},
  a2c2/.style={m, cPriv!75, densely dashed, mark=triangle*},
  frtc/.style={m, cPriv, densely dashed, mark=square*},
  deflect/.style={m, cPriv!70!black, densely dashed, mark=diamond*},
  deflectL/.style={m, cPriv!70!black, mark=diamond*},
  frtcL/.style={m, cPriv, mark=square*},
  strap/.style={cStrap, line width=1.7pt, mark=*, mark size=1.5pt},
  syncref/.style={black!60, densely dotted, line width=0.6pt, no marks},
}
\begin{figure}[t]
  \centering
  \begin{tikzpicture}
    \begin{axis}[delayplot, name=lib,
        title={LIBERO, SmolVLA}, xlabel={delay $K$}, ylabel={success (\%)},
        xmin=1, xmax=10, xtick={1,2,4,6,8,10}, ymin=68, ymax=86, ytick={70,75,80,85}]
      \addplot[syncref, domain=1:10] {83.35};
      \node[font=\tiny, black!60, anchor=south east] at (axis cs:10,83.45) {synchronous};
      \addplot[naive] coordinates {(1,82.10)(2,80.70)(3,81.05)(4,78.90)(5,77.70)(6,77.00)(7,75.10)(8,73.95)(9,72.85)(10,70.40)};
      \addplot[rtc] coordinates {(1,81.70)(2,81.95)(3,82.25)(4,79.55)(5,77.40)(6,76.95)(7,75.55)(8,73.95)(9,72.80)(10,70.90)};
      \addplot[trtc] coordinates {(1,80.90)(2,78.75)(3,78.70)(4,78.30)(5,77.25)(6,77.30)(7,77.00)(8,75.10)(9,74.90)(10,75.05)};
      \addplot[vlash] coordinates {(1,78.40)(2,76.10)(3,76.35)(4,75.55)(5,75.55)(6,74.50)(7,73.35)(8,72.95)(9,71.60)(10,70.80)};
      \addplot[deflectL] coordinates {(1,81.70)(2,81.70)(3,81.30)(4,79.40)(5,79.05)(6,76.40)(7,75.80)(8,72.55)(9,73.15)(10,70.55)};
      \addplot[frtcL] coordinates {(1,82.80)(2,81.60)(3,79.75)(4,79.25)(5,78.55)(6,77.10)(7,77.55)(8,76.60)(9,76.55)(10,75.50)};
      \addplot[strap] coordinates {(1,83.15)(2,82.05)(3,82.30)(4,82.35)(5,81.55)(6,80.05)(7,79.90)(8,79.40)(9,77.70)(10,78.70)};
    \end{axis}
    \begin{axis}[delayplot, at={(lib.east)}, anchor=west, xshift=0.75cm,
        title={Kinetix}, xlabel={delay $K$},
        xmin=0, xmax=4, xtick={0,1,2,3,4}, ymin=45, ymax=95, ytick={50,60,70,80,90},
        legend to name=delaylegend,
        legend columns=5, legend cell align=left,
        legend style={font=\tiny, draw=none, column sep=2pt, row sep=-1.5pt},
        legend image code/.code={\draw[mark repeat=2, mark phase=2] plot coordinates {(0cm,0cm) (0.17cm,0cm) (0.34cm,0cm)};}]
      \addplot[syncref, domain=0:4, forget plot] {91.5};
      \addplot[naive] coordinates {(0,91.5)(1,76.4)(2,70.7)(3,58.2)(4,49.0)}; \addlegendentry{Naive async}
      \addplot[te] coordinates {(0,74.2)(1,63.7)(2,57.4)(3,52.9)(4,52.1)}; \addlegendentry{TE}
      \addplot[bid] coordinates {(0,91.6)(1,87.6)(2,81.2)(3,68.4)(4,55.5)}; \addlegendentry{BID}
      \addplot[rtch] coordinates {(0,91.4)(1,84.6)(2,81.4)(3,73.1)(4,61.2)}; \addlegendentry{RTC (hard)}
      \addplot[rtc] coordinates {(0,92.1)(1,89.3)(2,84.4)(3,74.7)(4,61.1)}; \addlegendentry{RTC}
      \addplot[trtc] coordinates {(0,89.0)(1,88.6)(2,86.4)(3,81.8)(4,75.5)}; \addlegendentry{T-RTC}
      \addplot[remac] coordinates {(0,89.9)(1,89.0)(2,86.2)(3,80.3)(4,73.9)}; \addlegendentry{REMAC}
      \addlegendimage{empty legend}\addlegendentry{}
      \addlegendimage{empty legend}\addlegendentry{}
      \addplot[vlash] coordinates {(0,90.1)(1,89.1)(2,87.6)(3,85.7)(4,81.5)}; \addlegendentry{VLASH}
      \addplot[deflect] coordinates {(0,90.7)(1,89.9)(2,88.7)(3,87.7)(4,84.6)}; \addlegendentry{DEFLECT}
      \addplot[frtc] coordinates {(0,91.4)(1,90.6)(2,89.5)(3,87.4)(4,85.9)}; \addlegendentry{FutureRTC}
      \addplot[strap] coordinates {(0,91.2)(1,90.8)(2,89.9)(3,88.7)(4,86.5)}; \addlegendentry{\textbf{DelayOPSD}}
    \end{axis}
    \node[anchor=north] at ($(current bounding box.south) + (0,-0.05)$) {\pgfplotslegendfromname{delaylegend}};
  \end{tikzpicture}
  \caption{\textbf{Success under increasing inference delay}, all methods of
  \cref{tab:smolvla-k,tab:kinetix}. Gray: inference-time. Orange:
  training-time. Purple dashed: privileged (\cref{tab:smolvla-k,tab:kinetix}, $^\dagger$). Purple solid:
  FutureRTC and DEFLECT (LIBERO). Dotted: synchronous.}
  \label{fig:delay}
  \vspace{-10pt}
\end{figure}
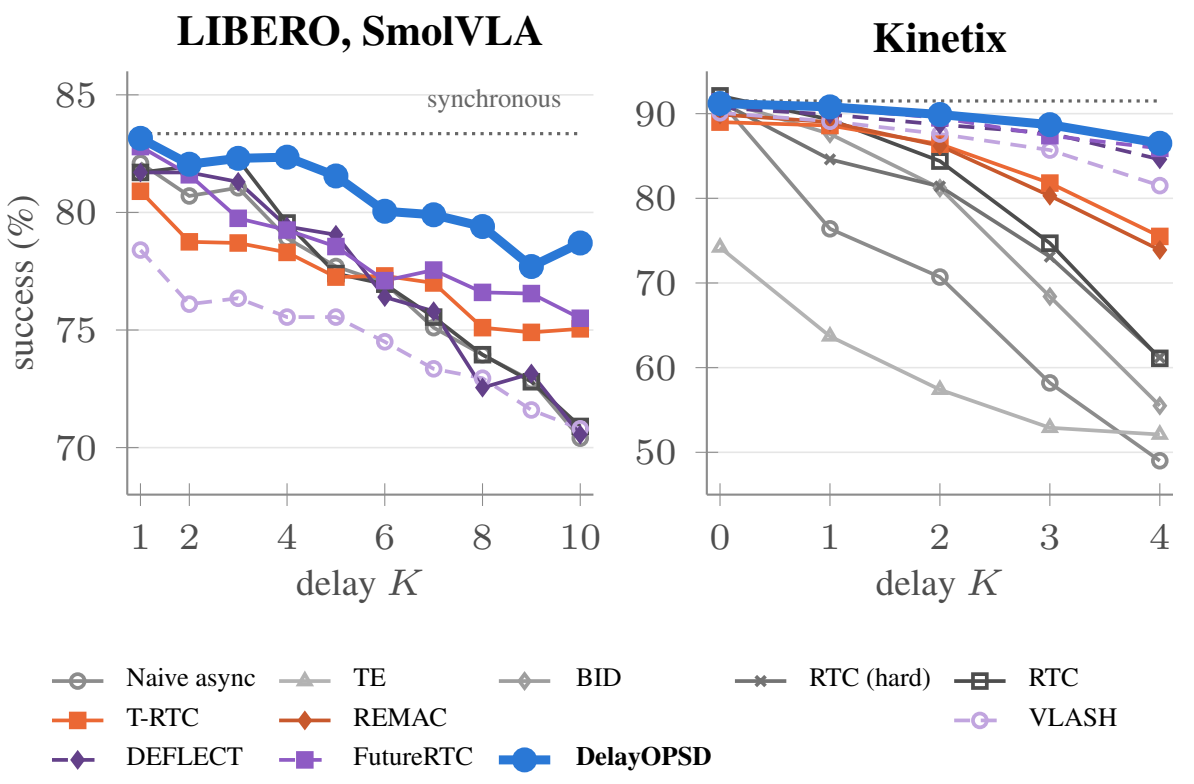

\begin{table}[t]
  \centering
  \footnotesize
  \setlength{\tabcolsep}{2.2pt}
  \caption{\textbf{Real robot.} $\pi_{0.5}$ on a YAM arm, $15{:}15$ at
  30\,Hz, 5 unseen scenes $\times$ 10 trials per task and method.
  Drawer/Basket: open a drawer, put a ball in the drawer/basket, close
  the drawer. Time: successful trials. Jump: median change of the commanded
  joint velocity at chunk switches.}
  \label{tab:real}
  \tabstyle
  \begin{tabular}{lccccc}
    \toprule
    & \multicolumn{3}{c}{Success (\%)} & & \\
    \cmidrule(lr){2-4}
    Method & Drawer & Basket & Avg. & Time (s) & Jump (rad/step) \\
    \midrule
    Synchronous & 66 & \textbf{52} & 59 & 40.7 & \textbf{0.028} \\
    Naive async & 50 & 42 & 46 & 30.4 & 0.079 \\
    RTC~\cite{rtc} & 52 & 46 & 49 & 30.0 & 0.053 \\
    \oursrow DelayOPSD & \textbf{74} & 50 & \textbf{62} & \textbf{29.5} & 0.030 \\
    \bottomrule
  \end{tabular}
  \vspace{-10pt}
\end{table}

\subsection{Setup}
\label{sec:exp-setup}

\paragraph{LIBERO.} We evaluate on the four LIBERO~\cite{libero} suites
(Spatial, Object, Goal, Long), each with 10 tasks, using 50 episodes per
task (500 per suite) from recorded initial states (\cref{sec:impl}). We fix $R{=}10$, the default of SmolVLA and $\pi_{0.5}$, and sweep
$K \le R$, with $K{=}0$ as the synchronous reference. Each entry averages 2,000
episodes (standard error 0.3--1.0 points, 95\% CIs in \cref{sec:ci}).

\paragraph{Training data.} For each delay, the buffer holds 1,600 episodes (40 per
task) on randomized scenes disjoint from evaluation, with one tuple per replanning step. The demonstration set has 1,693 episodes. Synchronous episodes on
the same scenes form the $K{=}0$ buffer. A single student is trained on the
buffers of all eleven delays (17,600 episodes) and serves every delay. After the scene filter, it trains on 253K of 340K tuples, against 273K
demonstration frames (\cref{tab:impl}). Baselines trained on demonstrations use no rollouts.

\paragraph{Kinetix.} We follow the RTC benchmark~\cite{rtc,kinetix}: 12
highly dynamic levels, simulated delays $K \in \{0, \dots, 4\}$,
$R = \max(K, 1)$, mean solved rate.

\paragraph{Real robot.} $\pi_{0.5}$ runs at 30\,Hz under $15{:}15$ execution. The delay of 0.5\,s equals the 10-step delay on LIBERO (20\,Hz) and exceeds the median latency of inference-time RTC (412\,ms). DelayOPSD learns from 100 delayed rollouts of 2 tasks,
without the scene filter, with a flow-matching term on demonstrations
for this small buffer (\cref{sec:impl-real}).

\paragraph{Baselines (\cref{tab:impl}).} \emph{Inference-time} methods are
naive asynchronous execution, temporal ensembling (TE)~\cite{act}, BID~\cite{bid}
and RTC~\cite{rtc}. Training-time RTC (T-RTC)~\cite{trainingtimertc} and
REMAC~\cite{remac} fine-tune with ground-truth action chunks as targets and run
with the committed actions as input. VLASH~\cite{vlash} takes the execution-time robot state as input. At evaluation, its released code reads this state from the simulator, and we mark methods that do so as \emph{privileged} ($\dagger$). On a real robot this state can only be approximated, for example by the last committed action. DEFLECT~\cite{deflect} fine-tunes on chunk pairs
from a frozen reference policy at demonstration timesteps. On SmolVLA and $\pi_{0.5}$, it
starts from the released LIBERO policy and takes no added input. On Kinetix, it builds on VLASH and takes the same simulator state
($\dagger$). FutureRTC~\cite{futurertc} keeps the policy frozen and adds the modules of \cref{sec:related}. Its released predictors were trained on demonstration and
rollout latents mixed 1:1, and in its released Kinetix code the robot state comes from the
simulator ($\dagger$).

\subsection{What each part contributes}
\label{sec:ablations}

\paragraph{Where the delay changes the actions.}\label{sec:diagnosis} The delay changes the policy's actions little on demonstrations but substantially on its own delayed rollouts. We measure the cost of the delay at a state as the difference between the
policy's actions given the inference-time observation $\vo_t$ and given the
execution-time observation $\vo_{t+K}$. At each recorded replanning step we run the
base on both without the prefix (\cref{tab:gap}).

\begin{table}[t]
  \centering
  \footnotesize
  \setlength{\tabcolsep}{4pt}
  \caption{\textbf{The cost of delay by state.}
  Squared difference ($\times 10^{-2}$) of mean executed actions, SmolVLA
  base at $\vo_{t+K}$ vs.\ at $\vo_t$ (no prefix),
  ${\sim}340$ contexts per cell (\cref{sec:gap-details}).}
  \label{tab:gap}
  \begin{tabular}{@{}lccc@{}}
    \toprule
    Contexts & $K{=}1$ & 5 & 10 \\
    \midrule
    Demonstrations & 0.23 & 0.53 & 0.51 \\
    \midrule
    Delayed rollouts, successes & 0.69 & 2.85 & 4.69 \\
    Delayed rollouts, failures & 3.38 & 16.82 & 37.21 \\
    \bottomrule
  \end{tabular}
  \vspace{-10pt}
\end{table}

On demonstrations, the difference stays below $0.6 \times 10^{-2}$ at every $K$. The base was trained on these chunks, so it reproduces the same demonstrated actions from $\vo_t$ and from $\vo_{t+K}$. On the delayed policy's own rollouts, the difference grows with $K$ and is largest in failed episodes. The delay also makes failures nearly twice as frequent (29.6\% vs.\ 16.7\% at
$K{=}10$).

\paragraph{The delayed policy's own states.} The teacher helps most on the delayed policy's own states. With the same teacher, SmolVLA at $K{=}10$ reaches 70.2\% when trained on demonstrations, 75.5\% on the base's synchronous rollouts and 78.7\% on its own delayed rollouts (\cref{tab:ablations}a, all CIs exclude zero). For $\pi_{0.5}$, the first and last are 90.3\% and 95.9\% (\cref{tab:ablations-k5}). Training-time RTC, the strongest method trained only on demonstrations, reaches 75.1\% (\cref{tab:smolvla-k}). Training on both the successes and the failures of the delayed rollouts works best (\cref{tab:ablations}b). For SmolVLA and $\pi_{0.5}$, training on successes alone gives 74.4\% and 95.5\%, on failures alone 66.3\% and 94.1\%, and on both with the scene filter 78.7\% and 95.9\%. Without a teacher, fine-tuning the delayed policy to imitate its own actions in successful rollouts gives 72.3\% (\cref{sec:selfbc}).

\paragraph{The teacher on failed states.} At failed states, the teacher departs from the previous plan and recovers more often than the delayed policy. We measure this with an \emph{old-plan index} on failed states held out from training. It is 0 for the teacher's chunk and 1 for the previous plan, and a higher value means that a new chunk stays closer to the previous plan (\cref{sec:copy-takeover}). At $K{=}10$, the index is 0.54 for the delayed base, 0.51 after training on demonstrations, 0.26 after training on synchronous rollouts and 0.21 for DelayOPSD. Training on rollouts thus moves the student toward the teacher's chunk, while training on demonstrations barely does. We then reset the simulator to these failed states and let each policy continue. In scenes where the synchronous policy succeeds, synchronous execution completes the task from 10.4\% of the failed states and the delayed base from 6.8\% ($+3.6$ $[1.9, 5.3]$). DelayOPSD completes it from 8.4\%, between the two and above the delayed base ($+1.6$ $[0.3, 3.1]$). In scenes where the synchronous policy fails, synchronous execution recovers from failed states about as rarely as the delayed base (4.3\% vs.\ 3.3\%). The scene filter drops failed episodes from these scenes.

\paragraph{Student or teacher samples.} In denoising time, matching velocities at the student's own samples works better than at re-noised teacher samples. \Cref{tab:ablations}a compares three choices of denoising states on delayed rollouts. Re-noised teacher samples add forward-process noise to the teacher's chunks, and the student matches the teacher's velocity at these noised chunks. Re-noised student samples add forward-process noise to the student's chunks instead, as DiffusionNFT~\cite{diffusionnft} does. The student path, used by DelayOPSD, keeps the intermediate states of the student's own sampler, as in on-policy distillation for diffusion and flow models~\cite{diffusionopd,flowopd}. At $K{=}10$, the three give 73.5\%, 77.0\% and 78.7\%. Averaged over $K{=}1$ to $10$, the student path is 3.4 $[2.5, 4.2]$ points above re-noised teacher samples, and re-noised student samples are 0.5 $[-0.2, 1.2]$ points below it (\cref{tab:denoising-delays}). Most of the gain over re-noised teacher samples therefore comes from matching velocities at the student's samples instead of the teacher's.

\paragraph{Committed actions.} Conditioning on the committed actions adds $2.6$ $[1.0, 4.2]$ points for SmolVLA and $4.2$ $[3.0, 5.5]$ for $\pi_{0.5}$ at $K{=}10$ (\cref{tab:ablations}c), and lowers the old-plan index of SmolVLA from 0.34 to 0.21.

\paragraph{One collection.} For both models, collecting the buffer during training or adding a second round of the trained student's rollouts gives no significant gain over collecting it once before training (\cref{tab:ablations}d). The real-robot buffer is also collected once.

\subsection{Comparison with prior methods}
\label{sec:main-results}

\paragraph{LIBERO, SmolVLA (\cref{tab:smolvla-k}).} At $K{=}10$, naive execution loses 13.0 points against synchronous execution, and DelayOPSD recovers 8.3 of these points (78.7\% vs.\ 70.4\%). For $K{\ge}4$, DelayOPSD is 1.2--3.2 points above FutureRTC. For $K{\le}3$, the best methods lie within 0.4 points of each other. Without the scene filter, DelayOPSD averages 79.4\% over $K{=}1$--10, above every baseline and 0.9 points above FutureRTC ($[-0.2, 1.9]$, \cref{tab:nofilter-delays}).

\paragraph{LIBERO, $\pi_{0.5}$ (\cref{tab:libero-main}).} At $K{=}10$, naive execution loses 11.1 points against the synchronous base, and DelayOPSD recovers 9.3 of these points. It is above FutureRTC by 0.6 $[0.1, 1.1]$ points on average and by 0.9 $[0.3, 1.5]$ at $K{\ge}5$. At $K{\le}4$, DelayOPSD and FutureRTC are both within noise of synchronous execution.

\paragraph{Kinetix (\cref{tab:kinetix}).} DelayOPSD is 5.1 points above training-time RTC, the best training-time baseline, on average and 11.0 points above it at $K{=}4$. Without simulator state, DelayOPSD matches or exceeds the privileged methods ($\dagger$).

\paragraph{Real robot (\cref{tab:real}).} DelayOPSD matches synchronous success while finishing faster. Naive execution lowers success
from 59\% to 46\%, and inference-time RTC reaches 49\%. DelayOPSD reaches 62\%, 16 points (95\% CI $[2, 30]$) above naive execution. It completes tasks in 28\% less time than synchronous execution, and its action jump at chunk switches is close to that of synchronous execution.
\section{Conclusion}
\label{sec:conclusion}

We found that the execution delay changes a flow-based VLA policy's actions far more on its own delayed rollouts than on demonstrations. Motivated by this, we proposed DelayOPSD, an on-policy self-distillation method in which the synchronous policy, given the execution-time observation, teaches the delayed policy on these rollouts and along the student's denoising trajectory. Without added inference-time modules, DelayOPSD achieves the highest average success among the compared methods. It needs no rollouts interleaved with training, which keeps training on a real robot practical. A limitation of DelayOPSD is that delayed rollouts of the base must be collected for every training delay (\cref{sec:limitations}). Reducing the number of rollouts and evaluating delays that vary within an episode are left for future work.
\section*{Acknowledgments}
LLM-based assistants, Claude (Anthropic) and ChatGPT (OpenAI), were used in the writing and in parts of the research.
In the writing, they edited and drafted parts of the text in all sections, summarized related work, gave reviewer-style feedback on drafts,
and helped produce the tables and figures (\LaTeX{} formatting and plotting scripts).
In the research, they helped write and debug the training, evaluation and analysis code, computed the confidence intervals,
checked the derivations in \cref{sec:method} and the appendix, and helped refine how the claims are stated.
The authors designed the method and experiments, ran all experiments, and
verified every statement, number and proof in the paper.

{
    \small
    \bibliographystyle{ieeenat_fullname}
    \bibliography{refs}
}

\clearpage
\renewcommand{\topfraction}{0.9}\renewcommand{\dbltopfraction}{0.9}\renewcommand{\textfraction}{0.08}\renewcommand{\floatpagefraction}{0.8}\renewcommand{\dblfloatpagefraction}{0.8}\setcounter{topnumber}{3}\setcounter{totalnumber}{4}
\setcounter{table}{0}\setcounter{figure}{0}\renewcommand{\thetable}{S\arabic{table}}\renewcommand{\thefigure}{S\arabic{figure}}

\appendix

\section{Implementation details}
\label{sec:impl}\label{sec:ci}

\Cref{tab:variants} gives the recipe of each setting and \cref{tab:cost} its cost. The stages are those of \cref{alg:strap}, and no model uses a separate training-time conditioning stage. At deployment DelayOPSD keeps the base's network and sampler, since the LoRA weights can be merged into the base.

\paragraph{Execution schedule.}\label{sec:impl-schedule} Under an $R{:}K$
schedule with $0 \le K \le R$ and $R + K \le H$, the replanning steps are
$t_j = jR$, $j \ge 1$. At $t{=}0$ the policy observes $\vo_0$ and the controller
waits for the first chunk, which executes its indices $[0, R+K)$ over steps
$[0, R+K)$. At each $t_j$ the policy observes $\vo_{t_j}$ and starts generating
while the running chunk continues. At $s_j = t_j + K$ the new chunk takes effect
and executes its indices $[K, K+R)$ over steps $[s_j, s_j + R)$. The prefix
$\vp_{t_j}$ holds the actions that the running chunk, generated at $t_{j-1}$
($t_0 = 0$), assigns to steps $t_j, \dots, t_j{+}K{-}1$, its indices
$[R, R{+}K)$. Other regimes, such as $K > R$ with several chunks in flight, are
out of scope.

\paragraph{Student and teacher inputs.} In \cref{eq:velocities}, the student's input chunk $[\vp_t; \va^{\tau}]$ has a per-token denoising time
$\bm{\tau} = [\vone_K; \tau \vone_{H-K}]$, so the prefix tokens are marked clean
and the suffix tokens carry the current denoising time. The teacher's $K$ masked tokens are zero-filled
and removed from the attention keys.
For example, with $H{=}50$ and $K{=}10$, the student's tokens 10 to 49 and the teacher's tokens 0 to 39 are both the actions of steps $t{+}10$ to $t{+}49$.
\emph{Suffix-masked flow matching}, which adapts the teacher where action tokens mix non-causally, is flow matching with the last $K$ input tokens
masked and excluded from the loss, for $K$ over the training delays.

\begin{table*}[tp]
  \centering
  \caption{\textbf{Recipe per setting.} Every student starts from the base, and every buffer holds delayed rollouts of the base under naive asynchronous execution. FM: flow matching. Details: \cref{sec:impl-smolvla,sec:impl-pi05,sec:impl-real,sec:impl-kinetix}.}
  \label{tab:variants}
  \footnotesize
  \setlength{\tabcolsep}{2.5pt}
  \begin{tabular}{@{}l*{4}{>{\raggedright\arraybackslash}p{0.212\linewidth}}@{}}
    \toprule
    & LIBERO, SmolVLA & LIBERO, $\pi_{0.5}$ & Real robot, $\pi_{0.5}$ & Kinetix \\
    \midrule
    Base & \texttt{lerobot/smolvla\_libero} & \texttt{pi05\_libero\_finetuned} & $\pi_{0.5}$ fine-tuned on 8 tasks & MLP-Mixer flow policy of RTC \\
    Schedule & $R{=}10$, $K{=}0, \dots, 10$, $H{=}50$, $N{=}10$ & $R{=}10$, $K{=}0, \dots, 10$, $H{=}50$, $N{=}10$ & $R{=}15$, $K{=}15$ (0.5\,s at 30\,Hz), $H{=}50$, $N{=}10$ & $R{=}\max(K,1)$, $K{=}0, \dots, 4$, $H{=}8$, $N{=}5$ \\
    Teacher & frozen base & base + suffix-masked FM & base + suffix-masked FM & base + suffix-masked FM \\
    Buffer & 1,600 episodes per delay, 17,600 for the eleven delays & 1,600 episodes per delay, 17,600 for the eleven delays & 100 episodes (2 tasks), one delay & $16 \times 2{,}048$ episodes per level and delay \\
    Failure rule & scene filter + stall cut & scene filter + stall cut & episode ends at a failure (\cref{sec:impl-real}) & time limit + stall cut \\
    Student & LoRA + action expert, 10K updates, batch 8 & LoRA + action expert, 20K updates, batch 4 & LoRA + action expert, 10K updates, batch 4, FM anchor on demonstrations & full network, 32 epochs, batch 512 \\
    Evaluation & 2,000 episodes per delay & 2,000 episodes per delay & 100 trials (2 tasks $\times$ 5 scenes $\times$ 10) & 2,048 episodes per level and delay, 12 levels \\
    \bottomrule
  \end{tabular}
\end{table*}

\subsection{LIBERO with SmolVLA}
\label{sec:impl-smolvla}

The base is \texttt{lerobot/smolvla\_libero}~\cite{smolvla}, and \cref{tab:impl} lists the settings and the baselines. Its action tokens attend causally, so the first $H{-}K$ tokens never attend to the
trailing $K$ and the teacher is the frozen base with no additional training. The $K{=}0$
tuples come from the synchronous buffer and have no prefix. Student and
teacher then receive the same input, so these tuples anchor the student to the
synchronous base.

\begin{table*}[tp]
  \centering
  \small
  \caption{\textbf{Settings on LIBERO.} Top: training and evaluation of DelayOPSD for SmolVLA (differences for $\pi_{0.5}$ in \cref{sec:impl-pi05}). Bottom: baselines for both models.}
  \label{tab:impl}
  \begin{tabular}{@{}lp{0.84\linewidth}@{}}
    \toprule
    Item & Setting \\
    \midrule
    \multicolumn{2}{@{}l}{\emph{DelayOPSD}} \\
    Buffers & Delayed rollouts of the base (naive $10{:}K$ execution) for $K{=}0, \dots, 10$ on 4 suites $\times$ 10 tasks,
      initial scenes disjoint from evaluation; 40 episodes per task
      and $K$ (1,600 per delay, 17{,}600 for the eleven delays);
      340K tuples, 253K after filtering (the demonstration ablation uses all 273K demonstration frames) \\
    Failed episodes & \emph{Scene filter:} a failed
      episode is dropped if its scene (same suite, task and initial state) has no success in the synchronous
      $10{:}0$ buffer, which removes 53.8\% of failed episodes for SmolVLA and 22.2\% for $\pi_{0.5}$.\newline \emph{Stall cut:} kept episodes are cut 30
      steps after the stall onset, the earliest step $i$ from which
      every end-effector position lies within 5\,cm of
      the mean position over $[i, T]$. The thresholds were fixed before training
      from the training buffers \\
    Flow step & $n$ uniform over $\{0, \dots, N-1\}$, pure-noise state
      included \\
    Trainable & LoRA~\cite{lora} on the VLM text model ($r{=}16$, $\alpha = 16$;
      q, k, v, o, gate, up, down); action expert, action/state projections and
      time MLP fully trained \\
    Initialization & LoRA $B{=}0$: the student starts with the teacher's
      parameters \\
    Optimization & 10K updates, batch 8, AdamW, lr $3\times10^{-5}$,
      500 warm-up updates, cosine decay to $2.5\times10^{-6}$, grad clip 10, fp32
      master weights; $\lambda = 30$ \\
    Student EMA & Decay 0.999; flow states use live
      weights; EMA weights are reported \\
    Evaluation & 4 suites $\times$ 10 tasks $\times$ 50 episodes, recorded
      initial states, seed 1000, MuJoCo 3.2.3; $10{:}K$ for $K{=}0, \dots, 10$ \\
    \midrule
    \multicolumn{2}{@{}p{0.98\linewidth}@{}}{\emph{Baselines.} Trained on the same delay range, $K \le 10$, with the
      authors' code where available and published hyperparameters otherwise.} \\
    Inf.-time RTC & Exponential guidance, $\beta = 5$, on the base \\
    Train.-time RTC & \emph{SmolVLA:} continues from
      the released checkpoint, full model, 30K updates, batch 32, lr
      $10^{-4}$, $K \sim \mathcal{U}\{0, \dots, 10\}$.\newline \emph{$\pi_{0.5}$:} the LeRobot implementation, full model from the LIBERO fine-tuned checkpoint, 30K updates, batch 8, lr $2.5\times10^{-5}$, $K \sim \mathcal{U}\{0, \dots, 10\}$, which gains 4.00 points at $K{=}10$
      (\cref{tab:libero-main}) \\
    VLASH & \emph{SmolVLA:} from the released
      checkpoint, full model, 30K updates, batch 32, lr $5\times10^{-5}$ as
      in its paper, with the fixed data loading of its code. Evaluated
      with the robot state at the step where the new chunk takes effect read from the simulator, as in
      its released evaluation.\newline \emph{$\pi_{0.5}$:} from \texttt{pi05\_base} as its LIBERO configuration, 50K updates, batch 8, $K \le 10$ \\
    DEFLECT & Our implementation of the paper (its code link has expired), pairs at recorded demonstration frames (Sec.~3.1 and App.~B of its arXiv v3).\newline \emph{SmolVLA:} one model per suite with the settings its paper reports for SmolVLA on LIBERO (its Tab.~3: 2,000 updates, batch 8, lr $10^{-6}$, action expert only), reference and initial policy the frozen \texttt{lerobot/smolvla\_libero}.\newline \emph{$\pi_{0.5}$:} the same recipe, reference and initial policy the frozen \texttt{pi05\_libero\_finetuned}, evaluated under naive asynchronous execution \\
    FutureRTC & Released checkpoints. Their stored training arguments list demonstration and rollout latent banks mixed 1:1 (rollout data not released) \\
    \bottomrule
  \end{tabular}
\end{table*}

\begin{table*}[tp]
  \centering
  \small
  \caption{\textbf{Training cost by stage} (GPU-h, on RTX 3090 for SmolVLA and Kinetix and on RTX 4090 for $\pi_{0.5}$). Episodes and updates are in \cref{tab:variants,tab:impl}. All stages run once, before
  deployment. Rollout collection includes the synchronous ($K{=}0$) buffer used by the scene filter.}
  \label{tab:cost}
  \begin{tabular}{@{}llll@{}}
    \toprule
    Stage & LIBERO, SmolVLA & LIBERO, $\pi_{0.5}$ & Kinetix \\
    \midrule
    Rollout collection & 11.4 (3.4M env.\ steps) & 18.5 (2.9M env.\ steps) & 7.7 (1.97M episodes, 503M env.\ steps) \\
    Teacher preparation & none (frozen base) & 3.1 & 1.5 \\
    Distillation & 6.6 & 16.9 & 5.2 \\
    \bottomrule
  \end{tabular}
\end{table*}

\subsection{LIBERO with $\pi_{0.5}$}
\label{sec:impl-pi05}

\paragraph{Teacher.} $\pi_{0.5}$'s action tokens are not causal, so the teacher is the base
fine-tuned with suffix-masked flow matching on the LIBERO demonstrations,
for $K{=}0, \dots, 10$ (action expert only, 10K updates, batch 8,
lr $10^{-5}$, exponential moving average (EMA) 0.999), then frozen. Run synchronously it reaches 97.70\%
(97.60\% with the last 10 actions masked), against 97.60\% for the base.

\paragraph{Student.} The buffers, the delays and the failure rule are those of SmolVLA (\cref{tab:impl}). The student uses LoRA ($r{=}16$, $\alpha{=}16$) on the PaliGemma
language model, trains the action expert and
the input, output and time projections fully, and keeps the vision encoder
frozen. The optimizer is AdamW, lr $2.5\times10^{-5}$ with 1,000 warm-up updates and cosine
decay to $2.5\times10^{-6}$, weight decay 0.01, grad clip 1.0, bf16,
20K updates, batch 4, on a single GPU.

\paragraph{VLASH over $K \le 10$.} VLASH executes a new chunk from its first action, so the chunk it must produce depends on the delay, which its input, the delayed image and the execution-time robot state, does not contain. Trained over all delays $K \le 10$, it loses success at every $K$: 87.15\% synchronously and 76.35\% at $K{=}10$, against 97.60\% for the synchronous base (\cref{tab:libero-main}). On SmolVLA it also loses synchronous success (77.15\% against 83.35\%), and on Kinetix, where $K \le 4$, it remains strong (\cref{tab:kinetix}). For VLASH, a loss at low delays when the fine-tuning delay range is widened is reported in~\cite{agouzoul2026async}, and lower success in a long-delay regime on LIBERO with $\pi_{0.5}$ in~\cite{eapn}. The other methods execute the part of the chunk that follows the delay, which is aligned in time for every $K$.

\paragraph{DEFLECT initialization.} For $\pi_{0.5}$, the DEFLECT paper starts from the VLASH checkpoint. Started from our VLASH checkpoint and evaluated as VLASH, DEFLECT averages 85.53\% over $K{=}1, \dots, 10$, at the level of that checkpoint (85.89\%). We therefore report it from the LIBERO fine-tuned base as for SmolVLA, where it averages 94.07\% against 93.15\% for the base under naive execution.

\paragraph{A2C2.} All methods are compared in the setting where each chunk is generated $K$ steps before its actions are executed. A2C2~\cite{a2c2} departs from this setting, since a lightweight correction head refines every executed action from the latest observation. We therefore exclude it from the comparison, consistent with prior asynchronous methods~\cite{vlash,deflect,futurertc}.

\subsection{Real robot with $\pi_{0.5}$}
\label{sec:impl-real}

\begin{figure*}[t]
  \centering
  \begin{tikzpicture}[lab/.style={font=\scriptsize,align=center,inner sep=1pt},ld/.style={draw=black!75,line width=0.4pt}]
    \node[anchor=south west,inner sep=0] (A) at (0,0) {\includegraphics[height=6.2cm]{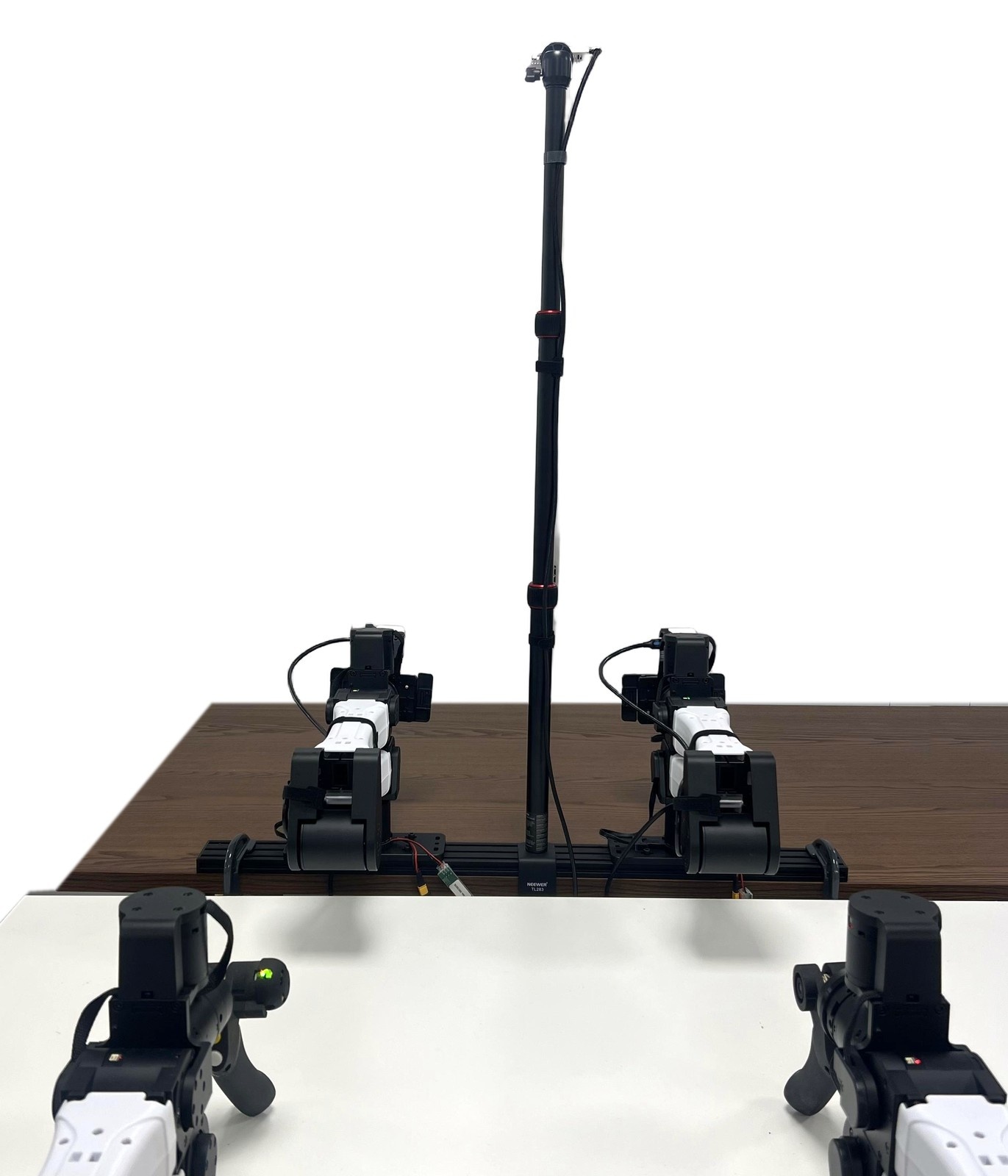}};
    \begin{scope}[x={(A.south east)},y={(A.north west)}]
      \node[lab,anchor=west] (tv) at (0.64,0.90) {Top-view camera};
      \draw[ld] (tv.west) -- (0.565,0.945);
      \node[lab,anchor=west] (wc) at (0.03,0.64) {Wrist camera};
      \draw[ld] (wc.south) -- (0.345,0.455);
      \node[lab,anchor=east] (fa) at (0.98,0.60) {Follower arm};
      \draw[ld] (fa.south) -- (0.715,0.39);
      \node[lab] (la) at (0.52,0.10) {Lead arms (teleoperation)};
      \draw[ld] (la.west) -- (0.23,0.13);
      \draw[ld] (la.east) -- (0.81,0.13);
      \node[font=\small,anchor=north west] at (0.0,1.0) {(a)};
    \end{scope}
    \node[anchor=south west,inner sep=0] (B) at ([xshift=2mm]A.south east) {\includegraphics[height=6.2cm]{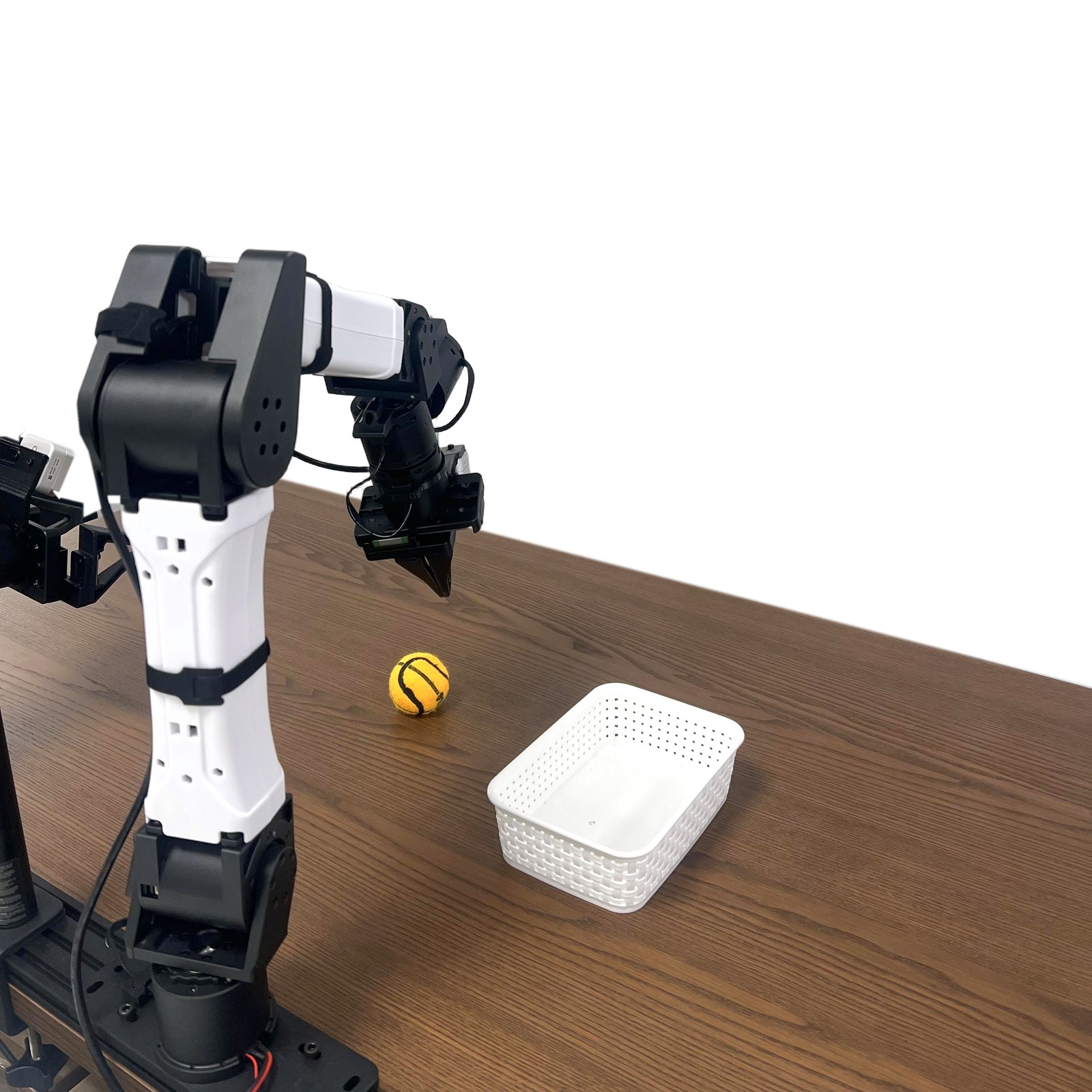}};
    \node[font=\small,anchor=north west] at (B.north west) {(b)};
    \node[anchor=north west,inner sep=0] (C0) at ([xshift=2mm]B.north east) {\includegraphics[height=2cm]{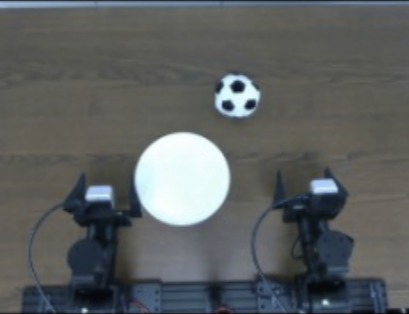}};
    \node[anchor=north west,inner sep=0] (C1) at ([yshift=-1mm]C0.south west) {\includegraphics[height=2cm]{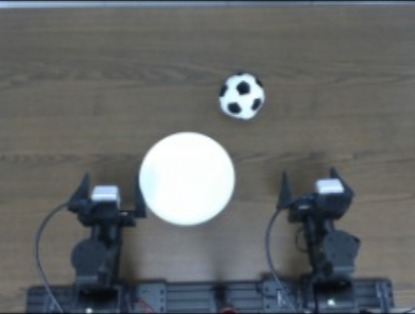}};
    \node[anchor=north west,inner sep=0] (C2) at ([yshift=-1mm]C1.south west) {\includegraphics[height=2cm]{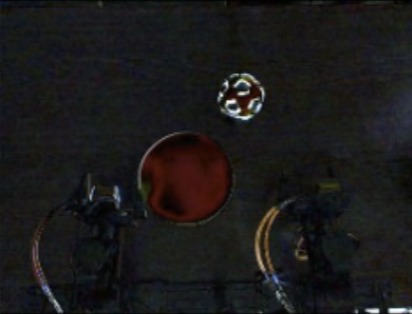}};
    \foreach \i/\t in {0/Reference,1/Current,2/Difference}{
      \node[font=\scriptsize,text=white,anchor=south west,inner sep=1.5pt] at (C\i.south west) {\t};
    }
    \node[font=\small,text=white,anchor=north west] at (C0.north west) {(c)};
  \end{tikzpicture}
  \caption{\textbf{Real-robot setup.} (a) Lead arms for teleoperation, the follower YAM arms, and the top-view and wrist cameras. (b) The follower arm. The objects belong to one of the demonstration tasks used to fine-tune the base, not to the evaluation tasks. (c) Object placement before each trial. The current top-view frame is compared with a reference frame of the scene, and their pixel difference guides the placement. The scene shown is an example.}
  \label{fig:real-setup}
\end{figure*}

\paragraph{Robot setup.} We use the right arm of a YAM robot with six joints and a parallel-jaw gripper, a top-view camera and a wrist camera, controlled at 30\,Hz (\cref{fig:real-setup}). $\pi_{0.5}$ is fine-tuned on demonstrations of 8 tasks with relative joint actions and predicts chunks of $H{=}50$ actions. This policy is the base of every method.

\paragraph{Policy execution.} All asynchronous methods run with the same delay of $K{=}15$ steps (0.5\,s), fixed before training above the median inference latency of every method (\cref{tab:real-exec}). The new chunk takes
over exactly at step $K$ with $R{=}15$. Shorter delays are evaluated on LIBERO (\cref{tab:smolvla-k,tab:libero-main}). Synchronous execution holds the arm during
inference. Inference-time RTC~\cite{rtc} runs on the same base policy with its authors' settings
(exponential schedule, maximum guidance weight 5). Of its 4,117 replies, 30 (0.7\%) arrived after step $K$, in which case
the running chunk continued until the reply arrived. All other replies of all methods arrived before step $K$.

\paragraph{Training.} The buffer holds 100 delayed rollouts of the base on 2 tasks
(50 each, 4,755 tuples), each on a different scene, with no synchronous runs
and no scene filter. An episode is ended as a failure when it exceeds one minute, when the robot pushes the drawer fully closed, or when a step is done out of order, such as grasping the ball before opening the drawer. Ending the episode there also removes the stalled tail that the stall cut removes on LIBERO.

The teacher is adapted with suffix-masked flow matching
for $K{=}15$ as for $\pi_{0.5}$ on LIBERO. Because this buffer is small, the
student's loss adds a flow-matching anchor on the two tasks' demonstrations (1:1) to guard against overfitting. The student trains for 10K updates with batch 4, and the rest of the
recipe is as for $\pi_{0.5}$ on LIBERO.

\paragraph{Evaluation.} The two tasks are Drawer (``open the drawer, pick up the basket ball, put it in the drawer, and close the drawer'') and Basket (``open the drawer, pick up the baseball ball, place it in the basket, and close the drawer''). Each method is evaluated on 5 scenes per task not used
in the buffer, with 10 trials per scene (\cref{tab:real-scenes}). Before each trial the objects are placed by hand to match a reference image of the scene from the top-view camera (\cref{fig:real-setup}c), so all methods start from approximately the same object positions. Completion time is the mean over successful trials of each task, averaged over the two tasks. The action jump
at a chunk switch is $\max_l |\Delta a_{t,l} - \Delta a_{t-1,l}|$ over the 6 arm joints $l$. Results are in \cref{sec:real-more}.

\subsection{Kinetix}
\label{sec:impl-kinetix}

On Kinetix we use the MLP-Mixer flow policies and benchmark code of
RTC~\cite{rtc}, not a VLA, with the objective of \cref{eq:opsd} and the recipe of \cref{tab:variants}. The MLP-Mixer mixes tokens with a
dense layer and cannot mask them, so the teacher is the base fine-tuned on expert data for 8 epochs with flow
matching plus suffix-masked flow matching, where a masked token is zero-filled at the input
and excluded from the loss. A level has a single initial scene, so the
scene filter does not apply. A failed episode keeps its tuples
while $t < 1.5\times$ the median goal-reaching length of its buffer and
before its stall onset (all dynamic bodies stay within a box of side 0.2).
A goal contact on any physics sub-step counts as success.

\paragraph{Baselines.} VLASH is re-evaluated from its released checkpoint (86.8\% on average, 86.7\% with the
environment's own seed for the simulator rollout of the committed actions). REMAC~\cite{remac} has no released checkpoint and is retrained from the base with
its released code and per-level learning rates for 32 epochs. DEFLECT has no released code or checkpoint
and is implemented from the paper on the released VLASH checkpoint (24 epochs, batch 512,
$\lambda_{\text{DPO}} = 0.02$, the paper's Kinetix value, $K \sim \mathcal{U}\{1, \dots, 4\}$), with the same simulator rollout as VLASH.

\subsection{Confidence intervals}
\label{sec:ci-sub}

Differences between two LIBERO models are paired by
task and episode index, and matched episodes share the seed and initial state. The 95\% confidence interval (CI) is a
percentile bootstrap over pairs (2,000 resamples, unstratified). The same initial scenes are evaluated at every delay, so pooled
CIs over several delays resample initial scenes, each with its pairs at all delays (10,000 resamples).
Against demonstrations, DelayOPSD's paired difference is $+4.15$ $[2.45, 5.90]$
at $K{=}5$ and $+8.55$ $[6.70, 10.35]$ at $K{=}10$. Against synchronous
rollouts, $+2.90$ $[1.25, 4.60]$ and $+3.25$ $[1.65, 4.95]$.
The pooled $\pi_{0.5}$ comparison with FutureRTC (\cref{sec:main-results}) uses the same initial-scene bootstrap.
The real-robot comparison (\cref{tab:real}) uses the normal-approximation CI for a difference of two success
rates, which is wider than the paired CI. Resampling the 10 scenes of \cref{tab:real-scenes} instead, each with the trials of both methods (stratified by task, 100{,}000 resamples), gives $+16$ $[4, 28]$ points for DelayOPSD against naive execution and $+13$ $[6, 21]$ against RTC.

\section{Empirical and theoretical analyses}
\label{sec:analysis}

\Cref{sec:gap-details,sec:selfbc,sec:copy-takeover} report measurements and ablations, and \cref{sec:loss-norm,sec:process,sec:rollout-proof,sec:euler-kl-proof,sec:return-proof,sec:kl-remarks,sec:decomp,sec:floor} contain the theory.

\subsection{Notation}
\label{sec:notation-sub}

\Cref{tab:notation} lists the notation. Throughout this section a replanning step $t$ is fixed unless an index $j$
is given. As in \cref{sec:pathloss}, $\va^{\tau}$ without a time range denotes the noisy suffix
$\va^{\tau}_{t+K:t+H}$. \Cref{eq:rollout} sums over $t = t_j$, $j = 1, \dots, M$.

\begin{table}[!h]
  \centering
  \small
  \setlength{\tabcolsep}{3pt}
  \caption{\textbf{Notation.} A bar denotes an average over the
  execution-time observation $\vo_{t+K}$ given $c_t$. The language instruction of a VLA is also an input and is omitted from the notation.}
  \label{tab:notation}
  \begin{tabular}{@{}>{\raggedright\arraybackslash}p{0.36\columnwidth}>{\raggedright\arraybackslash}p{0.60\columnwidth}@{}}
    \toprule
    Symbol & Meaning \\
    \midrule
    $R$, $K$, $H$ & replanning interval, delay, chunk length \\
    $\mathcal{K}$ & set of training delays \\
    $t_j = jR$, $s_j = t_j + K$ & $j$-th replanning step and the step at which its chunk takes effect \\
    $M$ & number of replanning steps \\
    $i$ & index in the suffix, for control step $t{+}K{+}i$ \\
    $n$, $\tau^n = n\Delta\tau$, $h = \Delta\tau = 1/N$ & denoising (Euler) step, its denoising time, step size \\
    $\vo_t$, $\vo_{t+K}$ & observation at the replanning step, and at the step where the new chunk takes effect \\
    $\vp_t$ & prefix: the $K$ committed actions for steps $t, \dots, t{+}K{-}1$ \\
    $c_t = (\vo_t, \vp_t)$ & information available to the student \\
    $\va^{\tau}$ & noisy suffix $\va^{\tau}_{t+K:t+H}$ ($\tau{=}0$ noise, $\tau{=}1$ clean action) \\
    $[\vp_t; \va^{\tau}]$, $\bm{\tau}$ & student's input chunk, with $\vp_t$ in place of the first $K$ noisy tokens, and its per-token denoising time \\
    $\va^{\tau^n}$, $\tilde\va^{\tau^n}$ & student and teacher denoising paths from the same noise \\
    $\vm$ & the $K$ masked tokens in the teacher's input \\
    $\theta$, $\theta_{\mathrm{base}}$, $\theta_T$ & student, base and teacher parameters \\
    $\mathcal{B}_K$, $\mathcal{B}$ & buffer of delayed rollouts of the base at delay $K$, and its union over $\mathcal{K}$ \\
    $\vv_S(\cdot \mid c_t)$, $\vv_T(\cdot \mid \vo_{t+K})$ & student and teacher velocity fields on the suffix \\
    $\bar\vv_T(\cdot \mid c_t)$ & averaged teacher velocity field \\
    $\pi_\theta(\cdot \mid c_t)$, $\pi_T(\cdot \mid \vo_{t+K})$ & student and teacher suffix distributions \\
    $P_\theta$, $P_*$ & trajectory distributions of the student and of $\pi^*$ \\
    $\ell(c_t, \vo_{t+K})$ & velocity loss along the student's denoising path \\
    $\bar\ell(c_t; \theta)$ & averaged velocity loss \\
    $\|\cdot\|_{\vw}$, $w_i = e^{-i/\lambda}$ & weighted norm and its weight on control step $t{+}K{+}i$ (\cref{sec:loss-norm}) \\
    $L$, $\rho = L/N$, $B$ & Lipschitz constant, Lipschitz ratio, score-gap constant \\
    \bottomrule
  \end{tabular}
\end{table}

\subsection{Teacher--policy differences and synchronous evaluation}
\label{sec:gap-details}
For \cref{tab:gap} the SmolVLA synchronous base acts twice at each logged
context: as the teacher at $\vo_{t+K}$, and as the delayed policy at $\vo_t$
without the prefix. We compare the executed actions $\va_{t+K:t+K+R}$ of the two, $R{=}10$, by the squared difference in normalized
action space averaged over action dimensions. Demonstration contexts come from the LIBERO training demonstrations of the base. Contexts are taken with the
buffer's stride of 10 from the four suites (${\sim}340$ per cell). Delayed
failures are those left by the failure rule of \cref{sec:physical}. The reported value removes sampling noise: from the
difference measured with independent initial noise for the two models (4
seeds each) we subtract each model's noise variance at its own context,
which leaves the squared difference of the mean outputs. At $K{=}10$ the 95\% CIs ($\times 10^{-2}$) are $[0.43, 0.62]$
(demonstrations), $[3.35, 6.38]$ (delayed successes) and $[31.4, 43.7]$
(delayed failures).

\paragraph{Synchronous evaluation.} We run the trained DelayOPSD SmolVLA checkpoint
synchronously ($10{:}0$, no prefix) on the scenes and initial states of the
synchronous base's evaluation, paired by episode:

\begin{center}
\footnotesize
\begin{tabular}{@{}lccccc@{}}
\toprule
 & Spatial & Object & Goal & Long & Avg. \\
\midrule
Synchronous base & 85.20 & 90.20 & 88.80 & 69.20 & 83.35 \\
DelayOPSD at $K{=}0$ & 84.80 & 91.00 & 86.00 & 69.60 & 82.85 \\
\bottomrule
\end{tabular}
\end{center}
The paired difference is $-0.50$ $[-2.05, 1.00]$ (2,000 paired episodes). DelayOPSD at $K{=}1$ (83.15\%)
is at the same level. Training on its own delayed rollouts, failures included, thus keeps the synchronous performance of the base, and the gain of DelayOPSD is specific to delayed execution.

\subsection{Ablations of training states and labels}
\label{sec:selfbc}
Without a teacher, the only labels on the delayed rollouts are the executed actions of successful episodes. Self-behavior cloning
(self-BC, \cref{tab:selfbc}) fine-tunes the delayed policy on them with the flow-matching loss, with the episodes and the student input of
the successes-only row of \cref{tab:ablations}b.

\begin{table}[h]
  \centering
  \small
  \caption{\textbf{Self-BC.} SmolVLA, LIBERO, four-suite average success
  (\%).}
  \label{tab:selfbc}
  \begin{tabular}{@{}llcc@{}}
    \toprule
    Episodes & Labels & $K{=}5$ & 10 \\
    \midrule
    Successes & Executed actions (self-BC) & 78.15 & 72.25 \\
    Successes & Teacher & 78.30 & 74.35 \\
    Filtered & Teacher (DelayOPSD) & \textbf{81.55} & \textbf{78.70} \\
    \bottomrule
  \end{tabular}
\end{table}

Self-BC cannot use the failed episodes, whose executed
actions are the behavior that failed, and the teacher labels them. With
the failures, DelayOPSD is $6.45$ $[4.70, 8.25]$ points above self-BC at
$K{=}10$. On the successful episodes alone, the teacher's labels are on par with the executed actions at $K{=}5$ and above them at $K{=}10$ ($+2.10$ $[0.20, 4.00]$). Self-BC improves on naive asynchronous execution of the base by 1.85
points at $K{=}10$.

\begin{table}[t]
  \centering
  \footnotesize
  \setlength{\tabcolsep}{1.8pt}
  \caption{\textbf{Physical states (columns) crossed with denoising states (rows).}
  LIBERO, four-suite average success (\%) at $K{=}5$ and $K{=}10$. Demonstrations: 273K frames, prefix from the
  demonstration actions. Bold: DelayOPSD. $^\ddagger$The training states of
  RFD~\cite{rfd}, our implementation, without its resolver.}
  \label{tab:crossing}
  \begin{tabular}{@{}lcccccc@{}}
    \toprule
    & \multicolumn{2}{c}{Demos} & \multicolumn{2}{c}{Sync.\ rollouts} & \multicolumn{2}{c}{Delayed rollouts} \\
    \cmidrule(lr){2-3}\cmidrule(lr){4-5}\cmidrule(l){6-7}
    Denoising states & $K{=}5$ & 10 & $K{=}5$ & 10 & $K{=}5$ & 10 \\
    \midrule
    \multicolumn{7}{@{}l}{\emph{SmolVLA}} \\
    Teacher, re-noised & 77.20$^\ddagger$ & 70.90$^\ddagger$ & 77.90 & 73.05 & 77.80 & 73.50 \\
    Student, re-noised & 76.70 & 70.05 & 78.80 & 74.65 & 81.35 & 76.95 \\
    Student path (ours) & 77.40 & 70.15 & 78.65 & 75.45 & \textbf{81.55} & \textbf{78.70} \\
    \midrule
    \multicolumn{7}{@{}l}{\emph{$\pi_{0.5}$}} \\
    Teacher, re-noised & 95.25$^\ddagger$ & 89.55$^\ddagger$ & 95.65 & 92.55 & 96.45 & 91.95 \\
    Student, re-noised & 95.75 & 90.20 & 96.20 & 93.55 & 96.90 & 95.90 \\
    Student path (ours) & 96.40 & 90.30 & 96.30 & 93.40 & \textbf{97.15} & \textbf{95.85} \\
    \bottomrule
  \end{tabular}
\end{table}

\begin{table}[t]
  \centering
  \small
  \setlength{\tabcolsep}{4pt}
  \caption{\textbf{Ablations of \cref{tab:ablations} at $K{=}5$ and $K{=}10$.} LIBERO, four-suite average success (\%). Each row
  changes one part of DelayOPSD. Physical states, episodes, input and collection as in
  \cref{tab:ablations}a--d. Two rounds: the trained student collects a second buffer and training continues on both.
  Online: the current student collects the tuples during training. Bold: DelayOPSD.}
  \label{tab:ablations-k5}
  \begin{tabular}{@{}lcccc@{}}
    \toprule
    & \multicolumn{2}{c}{SmolVLA} & \multicolumn{2}{c}{$\pi_{0.5}$} \\
    \cmidrule(lr){2-3}\cmidrule(l){4-5}
    Variant & $K{=}5$ & 10 & $K{=}5$ & 10 \\
    \midrule
    \multicolumn{5}{@{}l}{\emph{Physical states}} \\
    Demonstrations & 77.40 & 70.15 & 96.40 & 90.30 \\
    \midrule
    \multicolumn{5}{@{}l}{\emph{Episodes kept}} \\
    Successes only & 78.30 & 74.35 & 96.65 & 95.45 \\
    Failures only & 77.65 & 66.30 & 97.15 & 94.05 \\
    All episodes (unfiltered) & 79.40 & 76.25 & 97.00 & 95.75 \\
    \midrule
    \multicolumn{5}{@{}l}{\emph{Student input}} \\
    $\vo_t$ only (no prefix) & 80.35 & 76.10 & 95.60 & 91.65 \\
    \midrule
    \multicolumn{5}{@{}l}{\emph{Collection}} \\
    Two rounds & 79.95 & 76.15 & 97.35 & 95.80 \\
    Online & 79.45 & 76.45 & 96.60 & 95.95 \\
    \midrule
    DelayOPSD & \textbf{81.55} & \textbf{78.70} & \textbf{97.15} & \textbf{95.85} \\
    \bottomrule
  \end{tabular}
\end{table}

\begin{table}[t]
  \centering
  \footnotesize
  \setlength{\tabcolsep}{3pt}
  \caption{\textbf{Teacher's or student's samples at every delay.} SmolVLA, LIBERO, four-suite
  average success (\%) under $10{:}K$. Differences to the
  student path (DelayOPSD) in parentheses. Pooled
  differences with 95\% CIs are over $K{=}1, \dots, 10$.}
  \label{tab:denoising-delays}
    \begin{tabular}{@{}lccc@{}}
    \toprule
    & \multicolumn{1}{c}{Teacher's samples} & \multicolumn{2}{c}{Student's samples} \\
    \cmidrule(lr){2-2}\cmidrule(l){3-4}
    $K$ & Teacher, re-noised & Student, re-noised & Student path \\
    \midrule
    1 & 81.70 ($-$1.45) & 83.35 (+0.20) & 83.15 \\
    2 & 80.00 ($-$2.05) & 82.10 (+0.05) & 82.05 \\
    3 & 79.25 ($-$3.05) & 81.70 ($-$0.60) & 82.30 \\
    4 & 77.95 ($-$4.40) & 81.10 ($-$1.25) & 82.35 \\
    5 & 77.80 ($-$3.75) & 81.35 ($-$0.20) & 81.55 \\
    6 & 77.10 ($-$2.95) & 79.45 ($-$0.60) & 80.05 \\
    7 & 76.55 ($-$3.35) & 78.75 ($-$1.15) & 79.90 \\
    8 & 75.20 ($-$4.20) & 79.15 ($-$0.25) & 79.40 \\
    9 & 74.50 ($-$3.20) & 78.45 (+0.75) & 77.70 \\
    10 & 73.50 ($-$5.20) & 76.95 ($-$1.75) & 78.70 \\
    \midrule
    Pooled & $-$3.36 $[-4.17, -2.54]$ & $-$0.48 $[-1.18, 0.22]$ & -- \\
    \bottomrule
  \end{tabular}
\end{table}

\begin{table}[t]
  \centering
  \footnotesize
  \setlength{\tabcolsep}{4pt}
  \caption{\textbf{FutureRTC and DelayOPSD without the scene filter.} SmolVLA, LIBERO,
  four-suite average success (\%) under $10{:}K$ on the same 500 episodes per suite.
  Differences to FutureRTC in parentheses. The mean difference is paired, with its 95\% CI. With the scene filter (\cref{tab:smolvla-k}), the same paired mean difference to FutureRTC is $+2.19$ $[1.16, 3.20]$.}
  \label{tab:nofilter-delays}
    \begin{tabular}{@{}lcc@{}}
    \toprule
    $K$ & FutureRTC & No filter \\
    \midrule
    1 & 82.80 & 82.35 ($-$0.45) \\
    2 & 81.60 & 81.95 (+0.35) \\
    3 & 79.75 & 81.35 (+1.60) \\
    4 & 79.25 & 80.55 (+1.30) \\
    5 & 78.55 & 79.40 (+0.85) \\
    6 & 77.10 & 78.50 (+1.40) \\
    7 & 77.55 & 78.65 (+1.10) \\
    8 & 76.60 & 77.95 (+1.35) \\
    9 & 76.55 & 76.95 (+0.40) \\
    10 & 75.50 & 76.25 (+0.75) \\
    \midrule
    Mean & 78.53 & 79.39 ($+$0.86 $[-0.17, 1.90]$) \\
    \bottomrule
  \end{tabular}
\end{table}

\paragraph{Training-state variants.}\label{sec:denoising-delays} \Cref{tab:crossing,tab:ablations-k5,tab:denoising-delays,tab:nofilter-delays} report the ablations of \cref{tab:ablations} at $K{=}5$, for both models and per delay. The variants of \cref{tab:ablations}a are defined as follows.

\paragraph{Re-noised samples.} At the student's context, the teacher (or the
student, without gradient) generates a clean suffix with its $N$-step Euler
sampler. It is re-noised by the interpolation of \cref{sec:prelim} to a flow
time $\tau^n$ on the same grid, with $n$ drawn uniformly as for the student path, and the
student's velocity there is matched to the teacher's with the loss of
\cref{eq:opsd}.
Score matching on re-noised
teacher samples bounds the forward direction, for a stochastic
sampler with the teacher's exact score~\cite{song2021mle}.

\paragraph{Teacher's synchronous rollouts.} In \cref{tab:ablations}a and \cref{tab:crossing}, these are
tuples $(\vo_t, \vp_t, \vo_{t+K})$ from the teacher's $10{:}0$ rollouts of the same
scenes, $\vp_t$ the $K$ actions it executed after $t$, with the
same number of tuples and updates as DelayOPSD. They are a control, the policy's own states without the delay (CIs in \cref{sec:ci-sub}).

\subsection{Old-plan index and takeover from failures}
\label{sec:copy-takeover}
Both use held-out SmolVLA rollouts of the base (seed 3000, 40 episodes per
task at $10{:}10$, $10{:}5$ and $10{:}0$ on the same scenes, none in a training buffer,
base success 76.6\%, 80.9\%, 85.9\%, so not comparable to the main
evaluation). A failure is in the filter when the $10{:}0$ run of its scene
succeeds.

\paragraph{Old-plan index.} At a replanning step the teacher chunk $\vmu_T$
(synchronous base at $\vo_{t+K}$), the old plan $\vmu_{\mathrm{old}}$ (the base
at $\vo_{t-R}$, the plan that produced the prefix)
and the student chunk $\vmu_S$ are each a mean over 8 noise samples, in executed
action space over the full overlap of control steps. The index is
$\kappa = \sum \langle \vmu_S^{(1)} - \vmu_T^{(1)}, \vmu_{\mathrm{old}}^{(2)} - \vmu_T^{(2)}\rangle / \sum \langle \vmu_{\mathrm{old}}^{(1)} - \vmu_T^{(1)},
\vmu_{\mathrm{old}}^{(2)} - \vmu_T^{(2)}\rangle$ with independent noise sets $1, 2$, so that sampling noise
cancels in expectation in both sums, over the 25\% of contexts with the largest
$\lVert \vmu_{\mathrm{old}} - \vmu_T\rVert$ (selected on a third set, 1,200 failure contexts per
cell, 300 kept, task-balanced, episode-bootstrap CIs). By construction $\kappa = 0$
for the teacher chunk and $1$ for the old plan. At $K{=}10$ on failures:

\begin{center}
\footnotesize
\begin{tabular}{@{}lc@{}}
\toprule
Policy & $\kappa$ \\
\midrule
Delayed base & 0.54 $[0.51, 0.56]$ \\
Demonstration-trained student & 0.51 $[0.49, 0.53]$ \\
DelayOPSD without prefix & 0.34 $[0.32, 0.36]$ \\
Synchronous-rollout student & 0.26 $[0.24, 0.29]$ \\
DelayOPSD & 0.21 $[0.18, 0.23]$ \\
\bottomrule
\end{tabular}
\end{center}
The order is the same on successes and at $K{=}5$ (demonstration-trained student 0.35, DelayOPSD 0.14). 

\paragraph{Takeover.} At $K{=}10$, each failed episode (222 in the filter,
153 outside) is replayed from its logged actions, and
the full simulator state is saved at 25\%, 50\% and 75\% of the episode.
From each saved state, every policy restarts with no prefix (3
seeds): the synchronous base, the base under naive asynchronous execution (the delayed base), DelayOPSD, and the
demonstration-trained student. All takeovers of an episode are cut to the
steps left after its 75\% point, so later points differ in state, not in
time.

The recovery rates of \cref{sec:ablations} are over the three points. At the 25\% point, the synchronous base completes the task from 24.77\% of the states, the delayed base from 14.41\% and DelayOPSD from 18.02\%, and recovery is concentrated at this point. The demonstration-trained student is at the level of the delayed base in the filter ($+0.55$ $[-0.40, 1.60]$). Outside the filter the synchronous base is $+1.02$ $[-0.36, 2.47]$ points over the delayed base. Because the filter is defined by
synchronous success on the scene, the comparison inside it measures the
teacher on the scenes where training uses it.

\subsection{Loss normalization}
\label{sec:loss-norm}

A training tuple is sampled by drawing $K$ uniformly, then a task uniformly
within $\mathcal{B}_K$, then a tuple from that task, so that training
averages the bound over tasks and delays. The weight on control step $t{+}K{+}i$ is $w_i = e^{-i/\lambda}$ with $\lambda = 30$. For delay $K$ let $\vw_K = [w_0; \dots; w_{H-K-1}]$, so that $\|\vw_K\|_1 = \sum_{i=0}^{H-K-1} w_i$ and $W = d_a\, \mathbb{E}_K \|\vw_K\|_1$, with $d_a$ the action dimension. The norm of \cref{eq:opsd} is $\|\vu\|_{\vw}^2 = W^{-1} \sum_{i=0}^{H-K-1} w_i \|\vu_{t+K+i}\|^2$, so with $\vd = \vv_S - \mathrm{sg}(\vv_T)$, \cref{eq:opsd} is $\mathbb{E}\big[\sum_i w_i \|\vd_{t+K+i}\|^2\big] / W$. The normalizer $W$ is shared by all delays, so every control step of every tuple carries its weight $w_i$ and no delay is reweighted. In a batch of tuples $b$ with delays $K_b$, training uses
\begin{equation}
  \hat{\mathcal{L}}
  = \frac{\sum_b \sum_{i=0}^{H-K_b-1} w_i\, \|\vd_{b,t+K_b+i}\|^2}{d_a \sum_b \|\vw_{K_b}\|_1},
\end{equation}
where each tuple $b$ uses one draw of $\veps$ and $n$. This ratio is a self-normalized estimate of \cref{eq:opsd}: divided by the batch size, its denominator tends to $W$, and it does not depend on $\theta$.

\subsection{Process and the synchronous return}
\label{sec:process}

\paragraph{Execution.} The schedule is that of \cref{sec:impl-schedule}, with $M$ replanning steps $t_j = jR$ whose chunks take effect at $s_j = t_j + K$. We write $z_j$ for the environment state at $s_j$. The environment is Markov: $z_{j+1}$ and the reward of the $j$-th window
depend only on $z_j$, the executed window and fresh randomness. If an episode ends before the $M$-th replanning step, it stays in an absorbing state
with zero reward and no further suffix is generated, so the corresponding terms of
\cref{eq:rollout} are zero.

\paragraph{Reference policy and $J^*$.} The reference policy $\pi^*$ is the teacher run with the same $R{:}K$ timing: at each replanning step it observes $\vo_{s_j}$, the observation at the step where the chunk takes effect, and chooses the actions of the window $[s_j, s_j + R)$. Under both policies the first chunk is drawn as in the student's
deployment, so $z_1$ has the same
distribution. $J^*$ is the expected return of this process. After the first chunk it is
synchronous $R{:}0$ execution of the teacher with chunks of $H - K$ tokens, so $J^*$
differs from the teacher's synchronous score only by the effect of the first
chunk. With a reward of one on success at the end of
the episode and zero otherwise, the return is the success rate.

\paragraph{Trajectory distributions.} A trajectory lists, for each replanning step $j$, the history $\mathcal{H}_j$, the observation $\vo_{s_j}$ and the generated suffix $\va_{s_j:t_j+H}$:
{\small
\begin{equation*}
  \big(\mathcal{H}_1,\vo_{s_1},\va_{s_1:t_1+H},\,\dots,\,\mathcal{H}_M,\vo_{s_M},\va_{s_M:t_M+H},\,\mathcal{H}_{M+1},\vo_{s_{M+1}}\big).
\end{equation*}
}%
Here $\mathcal{H}_j$ is everything up to $s_j$ except $\vo_{s_j}$: environment states, observations, executed actions and
the first chunk. The student's context $c_{t_j}$ is a function of $\mathcal{H}_j$. Only the window $\va_{s_j:s_j+R}$ of each suffix is executed, and since $K \le R$ the next prefix
lies inside it. $P_\theta$ is the distribution of the trajectory under
the student's deployment and $P_*$ its distribution under $\pi^*$. The two distributions share
the distribution of $\mathcal{H}_1$, the kernel $\mathcal{H}_j \mapsto
\vo_{s_j}$, and the kernel $(\mathcal{H}_j, \vo_{s_j}, \va_{s_j:t_j+H}) \mapsto \mathcal{H}_{j+1}$,
which depends on the suffix only through the executed window. They differ only in how each suffix is drawn. The trajectory records the whole suffix, and the KL between the executed parts alone is no larger, by the data-processing inequality.

\subsection{Proof of \cref{thm:rollout}}
\label{sec:rollout-proof}

The proof applies the chain rule of the KL divergence to the process of
\cref{sec:process}. The step that requires care is that each factor of the chain rule is the student's suffix distribution $\pi_\theta(\cdot \mid c_{t_j})$, which holds because the student generates its suffix from $c_{t_j}$ without access to $\vo_{s_j}$. We first show this
(\cref{lem:suffix-indep}).

\begin{lemma}[The suffix is independent of $\vo_{s_j}$]
\label{lem:suffix-indep}
For the process of \cref{sec:process}, if the student's sampler noise is independent of the environment, then $\va_{s_j:t_j+H} \perp \vo_{s_j} \mid \mathcal{H}_j$, and under $P_\theta$
\begin{equation}
\begin{aligned}
  P_\theta(d\mathcal{H}_j, d\vo_{s_j}, d\va_{s_j:t_j+H}) ={}& P_\theta(d\mathcal{H}_j)\,
  \mathcal{T}(d\vo_{s_j} \mid \mathcal{H}_j) \\ &\times \pi_\theta(d\va_{s_j:t_j+H} \mid c_{t_j}),
\end{aligned}
\label{eq:suffix-factor}
\end{equation}
with $\mathcal{T}$ the environment's kernel. Under $P_*$ the same holds with $\pi_T(d\va_{s_j:t_j+H} \mid \vo_{s_j})$.
\end{lemma}

\begin{proof}
The student's suffix is $\va_{s_j:t_j+H} = F_\theta(c_{t_j}, \veps_j)$ with
$\veps_j \sim \mathcal{N}(\vzero, \rmI)$ independent of
$(\mathcal{H}_j, \vo_{s_j})$, and the prefix executed during $[t_j, s_j)$ was
committed before $t_j$. Hence $\va_{s_j:t_j+H} \perp \vo_{s_j} \mid \mathcal{H}_j$,
which gives \cref{eq:suffix-factor}. Under $P_*$ the suffix factor is
$\pi_T(\cdot \mid \vo_{s_j})$, since the synchronous teacher observes $\vo_{s_j}$
before acting. In simulation, time is frozen during inference and the
condition holds exactly.
\end{proof}

Following \cref{lem:suffix-indep} and the shared kernels of \cref{sec:process},
we now prove \cref{thm:rollout}.

\begin{restated}[\Cref{thm:rollout}]
For the process of \cref{sec:process}, if the student's sampler noise is
independent of the environment and both suffix distributions have positive
densities, then
\begin{equation*}
  \mathrm{KL}(P_\theta \,\|\, P_*) = \sum_{j=1}^{M}
  \mathbb{E}_{P_\theta}\Big[\mathrm{KL}\big(\pi_\theta(\cdot \mid c_{t_j})
  \,\big\|\, \pi_T(\cdot \mid \vo_{s_j})\big)\Big].
\end{equation*}
\end{restated}

\begin{proof}
By \cref{lem:suffix-indep} and the shared kernels of \cref{sec:process}, the
two trajectory distributions differ only in the suffix factor of each
replanning step, $\pi_\theta(\cdot \mid c_{t_j})$ under $P_\theta$ and
$\pi_T(\cdot \mid \vo_{s_j})$ under $P_*$, and both factors have positive
densities. The shared factors cancel in the density ratio, so
\begin{equation*}
  \log \frac{dP_\theta}{dP_*}
  = \sum_{j=1}^{M} \log \frac{\pi_\theta(\va_{s_j:t_j+H} \mid c_{t_j})}
  {\pi_T(\va_{s_j:t_j+H} \mid \vo_{s_j})} .
\end{equation*}
Taking the expectation under $P_\theta$ and conditioning each term on
$(\mathcal{H}_j, \vo_{s_j})$, under which the suffix is drawn from
$\pi_\theta(\cdot \mid c_{t_j})$ by \cref{lem:suffix-indep},
\begin{align*}
  \mathrm{KL}&(P_\theta \,\|\, P_*) \\
  &= \sum_{j=1}^{M} \mathbb{E}_{P_\theta}\bigg[\mathbb{E}\Big[\log
  \frac{\pi_\theta(\va_{s_j:t_j+H} \mid c_{t_j})}{\pi_T(\va_{s_j:t_j+H} \mid \vo_{s_j})}
  \,\Big|\, \mathcal{H}_j, \vo_{s_j}\Big]\bigg] \\
  &= \sum_{j=1}^{M} \mathbb{E}_{P_\theta}\Big[\mathrm{KL}\big(\pi_\theta(\cdot \mid c_{t_j})
  \,\big\|\, \pi_T(\cdot \mid \vo_{s_j})\big)\Big],
\end{align*}
which is \cref{eq:rollout}. Every term is nonnegative, so the sum is well
defined in $[0, \infty]$.
\end{proof}

\paragraph{Consequence for training.} The expectation in \cref{eq:rollout} is
under $P_\theta$, the student's own delayed rollouts. This follows from the
choice of the reverse direction for \cref{thm:rollout}, and DelayOPSD
therefore trains on delayed rollouts (\cref{sec:physical}). They are collected once with the base, and \cref{sec:kl-remarks} compares them with those of the trained student. Since
\cref{eq:rollout} is an equality, both the upper bound of
\cref{sec:return-proof} and the information floor of \cref{sec:floor} hold
for the trajectory KL.

\subsection{Proof of \cref{thm:euler}}
\label{sec:euler-kl-proof}

\paragraph{Idea of the proof.} Both
distributions are endpoints of $N$-step Euler chains from the same noise: the student's, with the
prefix clamped, and the teacher's first $H-K$ tokens, shifted by
$K$ and run on the later observation $\vo_{t+K}$. \Cref{lem:interp} extends each Euler chain to a continuous flow whose velocity error is controlled by $\ell$. For two flows, the KL changes at a rate equal to the product of their velocity difference and their score difference (\cref{lem:kl-flow}, from~\cite{su2025flowkl}). Integrating this rate, the Cauchy--Schwarz inequality gives the square root. We first state the assumptions and the two lemmas, and then prove \cref{thm:euler}.

\pagebreak
\begin{assumption}[Closed suffix]
\label{asm:closed}
The teacher's first $H-K$ tokens do not depend on the trailing $K$ (\cref{sec:impl-schedule}).
\end{assumption}

\begin{assumption}[Smoothness]
\label{asm:smooth}
$\vv_S$ and $\vv_T$ are $C^2$ and $L$-Lipschitz, with $\rho = L/N < 1$, where $h = 1/N$ is the Euler step.
\end{assumption}

\paragraph{Why the suffix must be closed.}\label{rem:closed} \Cref{thm:euler} compares the suffix distributions of the student and the teacher through their velocities on the suffix. The suffix loss controls the suffix distribution only if the suffix
dynamics do not depend on the discarded tokens. In two dimensions with
independent standard normal starts, $\vv_S(\xi_1, \xi_2) = (1, \xi_1)$ and
$\vv_T(\xi_1, \xi_2) = (0, \xi_1)$ have the same second component at every state, yet the
second coordinates end at $\mathcal{N}(1/2, 2)$ and $\mathcal{N}(0, 2)$, a KL
of $1/16$. This is why the teacher is causal over action tokens
(SmolVLA) or adapted with the trailing $K$ tokens masked ($\pi_{0.5}$,
Kinetix) (\cref{sec:impl-schedule}). On the student side the prefix is clean and
fixed, a condition rather than a generated coordinate.

\paragraph{Setup.} \emph{Coordinates.} Let $\mLambda$ be the diagonal matrix with entry
$w_i / W$ on every coordinate of the action at step $t{+}K{+}i$. After the change of variables $\va^{\tau} \mapsto \mLambda^{1/2} \va^{\tau}$, the weighted norm $\|\cdot\|_{\vw}$ is the Euclidean norm. The KL divergence is unchanged by this bijection, which is applied to both chains, and the initial noise becomes
$\mathcal{N}(\vzero, \mLambda)$. We work in these coordinates and keep the symbols, so gradients and scores below are taken in them.

\emph{The teacher's chain.} By Assumption~\ref{asm:closed}, the
teacher's first $H-K$ tokens form an Euler chain of their own, written $\tilde\va^{\tau^n}$, and
$\pi_T(\cdot \mid \vo_{t+K})$ is its endpoint distribution.

\emph{Notation.} Fix $(c_t, \vo_{t+K})$ and write $\vv_S$ and $\vv_T$ for $\vv_S(\cdot \mid c_t)$ and $\vv_T(\cdot \mid \vo_{t+K})$. Let
$\bm{\delta}_n(\vxi) = \vv_S(\vxi, \tau^n) - \vv_T(\vxi, \tau^n)$ be the difference of the two velocities at the same state $\vxi$. Then
$\ell(c_t, \vo_{t+K}) = h \sum_n \mathbb{E}\|\bm{\delta}_n(\va^{\tau^n})\|^2$, with the expectation over the student's chain. Within the $n$-th Euler step the denoising time is
$\tau = \tau^n + r$ with $r \in [0, h]$.

\begin{lemma}[Euler chains as flows]
\label{lem:interp}
For $k \in \{S, T\}$, a step $n$ and $r \in [0, h]$, let $G_{k,n,r}(\vxi) = \vxi + r\, \vv_k(\vxi, \tau^n)$ be a partial Euler step. Under Assumption~\ref{asm:smooth}:
\begin{list}{}{\setlength{\leftmargin}{2.6em}\setlength{\labelwidth}{2.0em}\setlength{\labelsep}{0.5em}\setlength{\itemsep}{0pt}\setlength{\topsep}{2pt}\setlength{\parsep}{0pt}}
  \item[(i)] $G_{k,n,r}$ is a $C^2$ diffeomorphism.
  \item[(ii)] The interpolations $\hat\va_S(\tau^n + r) = G_{S,n,r}(\va^{\tau^n})$ and $\hat\va_T(\tau^n + r) = G_{T,n,r}(\tilde\va^{\tau^n})$ pass through the Euler chains at $r \in \{0, h\}$ and
follow the velocity fields $\hat\vv_k(\vzeta, \tau^n + r) = \vv_k(G_{k,n,r}^{-1}(\vzeta), \tau^n)$. Their
distributions, $\hat p_\tau$ for the student and $\hat q_\tau$ for the teacher,
solve the continuity equation on each interval and are continuous in $\tau$ at the
nodes.
  \item[(iii)] $\|\hat\vv_S(\vzeta, \tau) - \hat\vv_T(\vzeta, \tau)\| \le \|\bm{\delta}_n(\vxi)\| / (1 - rL)$
with $\vxi = G_{S,n,r}^{-1}(\vzeta)$.
  \item[(iv)] $\hat{\mathcal{E}} := \int_0^1 \mathbb{E}_{\hat p_\tau}\|\hat\vv_S - \hat\vv_T\|^2 d\tau \le \ell(c_t, \vo_{t+K}) / (1 - \rho)$.
\end{list}
\end{lemma}

\begin{proof}
We drop the indices $k, n, r$ where they are clear.

(i) Since $\vv_k$ is $L$-Lipschitz and $rL \le \rho < 1$, the map $G = \mathrm{id} + r\, \vv_k$ is a bijection with $\|G(\vxi) - G(\vxi')\| \ge (1 - rL)\|\vxi - \vxi'\|$ and an invertible Jacobian, so $G^{-1}$ is $C^2$ by the inverse function theorem.

(ii) $\frac{d}{dr} G_{k,n,r}(\vxi) = \vv_k(\vxi, \tau^n) = \hat\vv_k(G_{k,n,r}(\vxi), \tau^n + r)$,
so each interpolated path is a characteristic of $\hat\vv_k$, and its distribution solves the continuity equation. At $r = h$ the
map is the full Euler step, and the next interval starts at $r = 0$, where the map is the identity.

(iii) Let $\vzeta = G_{S,n,r}(\vxi)$ and $\vxi' = G_{T,n,r}^{-1}(\vzeta)$. Both $\vxi$ and $\vxi'$ are mapped to $\vzeta$, so $\vxi' - \vxi = r\,(\vv_S(\vxi)
- \vv_T(\vxi')) = r\,\ve$, where $\ve = \hat\vv_S(\vzeta) - \hat\vv_T(\vzeta) = \bm{\delta}_n(\vxi) + \vv_T(\vxi) -
\vv_T(\vxi')$. Hence $\|\ve\| \le \|\bm{\delta}_n(\vxi)\| + L\|\vxi - \vxi'\| = \|\bm{\delta}_n(\vxi)\| + rL\|\ve\|$, which is the claim.

(iv) Under $\hat p_{\tau^n + r}$, the point $\vxi$ of (iii) is $\va^{\tau^n}$. Squaring (iii) and integrating over $r$,
\begin{align*}
  \hat{\mathcal{E}} &\le \sum_n \mathbb{E}\|\bm{\delta}_n(\va^{\tau^n})\|^2 \int_0^h
  \frac{dr}{(1 - rL)^{2}} \\ &= \sum_n \frac{h}{1 - hL}\, \mathbb{E}\|\bm{\delta}_n(\va^{\tau^n})\|^2
  = \frac{\ell(c_t, \vo_{t+K})}{1 - \rho} ,
\end{align*}
since $hL = \rho$.
\end{proof}

\begin{lemma}[KL between two flows, {\cite[Lemma 3.1]{su2025flowkl}}]
\label{lem:kl-flow}
Let $p_\tau, q_\tau$ be positive $C^1$ densities solving
$\partial_\tau p + \nabla \cdot (p\, \vf_1) = 0$ and
$\partial_\tau q + \nabla \cdot (q\, \vf_2) = 0$, with enough integrability that
the boundary terms below vanish. With $\bm{\omega}_1 = \nabla \log p_\tau$ and
$\bm{\omega}_2 = \nabla \log q_\tau$,
$\frac{d}{d\tau}\mathrm{KL}(p_\tau \,\|\, q_\tau) = \mathbb{E}_{p_\tau}\langle \vf_1 - \vf_2,\,
\bm{\omega}_1 - \bm{\omega}_2 \rangle$.
\end{lemma}

\paragraph{Score gap.} For the flows $\hat p_\tau$ and $\hat q_\tau$ of \cref{lem:interp}, let
\begin{equation}
  \mathcal{I}(c_t, \vo_{t+K}) = \int_0^1 \mathbb{E}_{\hat p_\tau}\big\|\nabla \log \hat p_\tau -
\nabla \log \hat q_\tau\big\|^2 d\tau .
  \label{eq:score-gap}
\end{equation}

\begin{assumption}[Score gap]
\label{asm:score}
The time-integrated squared score gap $\mathcal{I}$ of the two flows, defined in \cref{eq:score-gap}, is at most $B$ on the contexts reached, and the two flows satisfy the integrability of \cref{lem:kl-flow}.
\end{assumption}

Following \cref{lem:interp,lem:kl-flow}, under Assumptions~\ref{asm:closed} to~\ref{asm:score}, we now prove \cref{thm:euler}.

\begin{restated}[\Cref{thm:euler}]
Let $\va^{\tau}$ be the student's denoising path ($N$ Euler steps from noise) and
$\ell(c_t, \vo_{t+K}) = \mathbb{E}_{\veps, \tau}\|\vv_S(\va^{\tau}, \tau \mid c_t) - \vv_T(\va^{\tau},
\tau \mid \vo_{t+K})\|_{\vw}^2$, $\tau \sim \mathcal{U}\{0, \Delta\tau, \ldots, 1-\Delta\tau\}$. Under Assumptions~\ref{asm:closed} to~\ref{asm:score},
\begin{equation*}
  \mathrm{KL}\big(\pi_\theta(\cdot \mid c_t) \,\big\|\, \pi_T(\cdot \mid \vo_{t+K})\big)
  \le \sqrt{B\, \ell(c_t, \vo_{t+K}) / (1 - \rho)} .
\end{equation*}
\end{restated}

\begin{proof}
Both flows start from the
same distribution, so the KL is zero at $\tau = 0$, and their endpoints are
$\pi_\theta(\cdot \mid c_t)$ and $\pi_T(\cdot \mid \vo_{t+K})$. \Cref{lem:kl-flow} applies on each interval, and the KL is continuous at the nodes. Integrating its rate from $\tau = 0$ to $1$,
\begin{align*}
  &\mathrm{KL}\big(\pi_\theta(\cdot \mid c_t) \,\|\, \pi_T(\cdot \mid \vo_{t+K})\big) \\
  &\quad= \int_0^1 \mathbb{E}_{\hat p_\tau}\big\langle \hat\vv_S -
\hat\vv_T,\; \nabla \log \hat p_\tau - \nabla \log \hat q_\tau\big\rangle\, d\tau \\
  &\quad\le \sqrt{\hat{\mathcal{E}}\; \mathcal{I}} \;\le\; \sqrt{B\, \ell(c_t, \vo_{t+K}) / (1-\rho)} ,
\end{align*}
by the Cauchy--Schwarz inequality, \cref{lem:interp}(iv) and Assumption~\ref{asm:score}. This is \cref{eq:euler-kl}.
\end{proof}

\paragraph{Consequence for training.} The bound is used to identify where the velocity error should be reduced. The rate in \cref{lem:kl-flow} is an expectation under the student's flow $\hat p_\tau$, which is why DelayOPSD evaluates the loss on the student's denoising path (\cref{sec:pathloss}). If both chains used Euler--Maruyama steps with a common noise level, the KL between the path distributions of the two chains would be exactly a $\tau$-weighted sum of the same velocity differences, which is the per-step objective of DiffusionOPD and Flow-OPD~\cite{diffusionopd,flowopd}.

\subsection{Proof of the success-gap bound}
\label{sec:return-proof}

The two theorems give a bound on the difference in success rate between the student and the reference policy. With $\nu_\theta$ the distribution of the student's contexts over the
replanning steps of its deployment and
$\bar\ell(c; \theta) = \mathbb{E}_{\vo_{t+K} \mid c}\, \ell(c, \vo_{t+K})$,
\begin{equation}
  \big|J^* - J(\pi_\theta)\big| \le \sqrt{\tfrac{M}{2}}\,
  \Big(\tfrac{B}{1-\rho}\, \mathbb{E}_{c \sim \nu_\theta}\, \bar\ell(c; \theta)\Big)^{1/4} .
  \label{eq:return}
\end{equation}

\paragraph{Proof.} Write $\ell_j = \ell(c_{t_j}, \vo_{s_j})$. Then
\begin{align*}
  \mathrm{KL}(P_\theta \,\|\, P_*)
  &= \sum_{j=1}^{M} \mathbb{E}_{P_\theta}\, \mathrm{KL}\big(\pi_\theta(\cdot \mid c_{t_j}) \,\|\, \pi_T(\cdot \mid \vo_{s_j})\big) \\
  &\le \sqrt{\tfrac{B}{1-\rho}}\; \sum_{j=1}^{M} \mathbb{E}_{P_\theta} \sqrt{\ell_j} \\
  &\le M \sqrt{\tfrac{B}{1-\rho}\; \mathbb{E}_{c \sim \nu_\theta}\, \bar\ell(c; \theta)} ,
\end{align*}
by \cref{thm:rollout}, \cref{thm:euler} and Jensen's inequality. The last step uses $\frac{1}{M}\sum_j \mathbb{E}\sqrt{\ell_j} \le \sqrt{\frac{1}{M}\sum_j
\mathbb{E}\,\ell_j}$ and $\frac{1}{M}\sum_j \mathbb{E}\,\ell_j = \mathbb{E}_{c \sim \nu_\theta}\bar\ell(c; \theta)$. Success is an event $\mathcal{S}$ of the
trajectory, so by Pinsker's inequality
\begin{align*}
  \big|J^* - J(\pi_\theta)\big| &= \big|P_*(\mathcal{S}) - P_\theta(\mathcal{S})\big| \le
\mathrm{TV}(P_\theta, P_*) \\ &\le \sqrt{\mathrm{KL}(P_\theta \,\|\, P_*)/2} ,
\end{align*}
which gives \cref{eq:return}.

Applied to each task and delay and averaged, the concavity of
$\xi \mapsto \xi^{1/4}$ gives the same bound for the benchmark average, with
$\mathbb{E}_{\nu_\theta}\bar\ell$ replaced by its average over tasks and
delays. Training averages over tasks and delays in the same way (\cref{sec:loss-norm}).

\subsection{From the bounds to the training loss}
\label{sec:kl-remarks}

\Cref{eq:return} is stated for the velocity loss at the student's own contexts. The training loss of \cref{eq:opsd} relates to it as follows.

\paragraph{The training update.}\label{sec:surrogate} The trajectory KL depends on $\theta$ through the suffix distribution at each context and through which contexts are visited. At each update the denoising states are sampled anew from the current student, and the gradient is taken only through the student's velocity at them, as in on-policy distillation of language models~\cite{gkd}.

\paragraph{Buffer contexts.} Training evaluates the loss on the buffer of the base's delayed rollouts (\cref{sec:physical}). Protocols that collect it again during training are compared in \cref{tab:ablations}d and \cref{tab:ablations-k5}. We also collected a second buffer with the trained student on the same scenes at $K{=}5$ and $10$ and measured, as in \cref{tab:gap}, the student's difference to the teacher on both. On the states the trained student visits, this difference matches the one on its training buffer, for successes and failures and for both models: the ratio is 0.94 to 1.12 for SmolVLA, with every 95\% CI containing 1, and every paired difference for $\pi_{0.5}$ has a 95\% CI containing zero.

\subsection{What the training loss estimates}
\label{sec:decomp}

Define the future-averaged teacher field
\begin{equation}
  \bar\vv_T(\va^{\tau}, \tau \mid c_t) = \mathbb{E}_{\vo_{t+K} \sim p(\cdot \mid c_t)}\big[\,\vv_T(\va^{\tau}, \tau \mid \vo_{t+K})\,\big].
\end{equation}
The noise that produces $\va^{\tau}$ is drawn independently of $\vo_{t+K}$, so $\va^{\tau} \perp \vo_{t+K} \mid
c_t$ on the states generated by $\theta^{-}$, a no-gradient copy of $\theta$, and for each state
\begin{equation}
\begin{aligned}
  \mathbb{E}\big[\|\vv_S - \vv_T\|_{\vw}^2 \mid c_t, \va^{\tau}, \tau\big]
  &= \underbrace{\|\vv_S - \bar\vv_T\|_{\vw}^2}_{\text{learnable}} \\
  &+ \underbrace{\mathbb{E}\big[\|\vv_T - \bar\vv_T\|_{\vw}^2 \mid c_t, \va^{\tau}, \tau\big]}_{\text{information gap}} .
\end{aligned}
  \label{eq:decomp}
\end{equation}
With the states fixed by $\theta^{-}$, only the first term depends on
$\theta$. The second is the part of the teacher's velocity that the
student's input does not determine. The first term is zero when $\vv_S = \bar\vv_T$, so the regression estimates the teacher's field averaged over the execution-time observations that are possible given $c_t$. 

\subsection{What the student cannot recover}
\label{sec:floor}

This subsection proves the two statements of \cref{sec:prefix} about the information floor.

\paragraph{The floor.} Fix a joint distribution of $(c_t, \vo_{t+K})$ and let $p_T(\cdot \mid \vo_{t+K})$ be the
density of $\pi_T(\cdot \mid \vo_{t+K})$. For any distribution $q$ of the student's
suffix at $c_t$, expanding the logarithm gives
\begin{equation}
\begin{aligned}
  \mathbb{E}_{\vo_{t+K} \mid c_t}\, \mathrm{KL}\big(q \,\|\, \pi_T(\cdot \mid \vo_{t+K})\big)
  &= \mathrm{KL}(q \,\|\, q^*) - \log Z(c_t) \\ &\ge -\log Z(c_t),
\end{aligned}
\end{equation}
with $q^* \propto \exp\{\mathbb{E}_{\vo_{t+K} \mid c_t} \log p_T(\cdot \mid \vo_{t+K})\}$ and
$Z(c_t)$ its normalizer, assumed positive. By Jensen's inequality $Z \le 1$, and $Z = 1$ only
when $\pi_T(\cdot \mid \vo_{t+K})$ does not depend on $\vo_{t+K}$ given $c_t$. In particular the floor is zero when $\vo_{t+K}$ is a function of $c_t$, as under deterministic dynamics with an observation that
determines the state: the student can then set $\vv_S = \vv_T(\cdot \mid \vo_{t+K}(c_t))$ and reproduce
the teacher's distribution from the same noise. The regression of \cref{eq:opsd} is minimized by the averaged velocity $\bar\vv_T$ of \cref{sec:decomp}, and $q^*$ is the optimum of the reverse KL that the bounds concern. The two coincide when $\vo_{t+K}$ is determined by $c_t$.

\paragraph{The prefix lowers the floor.} Let $\Phi_1 = \mathbb{E}_{c_t}[-\log Z(c_t)]$ be the expected floor of a student that sees $c_t = (\vo_t, \vp_t)$, with optimal distribution $q^*_1(\cdot \mid c_t)$, and let $\Phi_0$ and $q^*_0(\cdot \mid \vo_t)$ be the same quantities for a student that sees only $\vo_t$. Such a student must use one distribution $q$ for all prefixes that can follow $\vo_t$. By the identity above at each $c_t$, its objective is
$\mathbb{E}_{\vp_t \mid \vo_t}\big[\mathrm{KL}(q \,\|\, q^*_1(\cdot \mid c_t)) - \log Z(c_t)\big]$. Taking the minimum over $q$, attained at $q^*_0$, and averaging over $\vo_t$ gives
\begin{equation}
  \Phi_0 - \Phi_1 = \mathbb{E}_{c_t}\, \mathrm{KL}\big(q^*_0(\cdot \mid \vo_t) \,\|\, q^*_1(\cdot \mid c_t)\big) \ge 0 .
  \label{eq:prefix-gain-kl}
\end{equation}
The committed actions thus lower the floor by how much they
change the optimal chunk distribution. Summed over replanning steps, $\mathrm{KL}(P_\theta \,\|\, P_*) \ge \sum_j \mathbb{E}_{P_\theta}
[-\log Z(c_{t_j})]$ for every $\theta$, with $Z$ computed under the
deployment's $\vo_{s_j} \mid c_{t_j}$. The floor concerns how closely the student can match the
teacher for each execution-time observation, not its success: actions that succeed under
every likely future may still exist.

\paragraph{Consequence for the method.} The student receives the committed actions as input (\cref{eq:pair}), and removing them lowers success (\cref{tab:ablations}c).

\section{Additional results}
\label{sec:suites}

\subsection{LIBERO speed and smoothness}
\label{sec:speed}

\Cref{tab:speed} gives the numbers behind \cref{fig:teaser}b. All rows use $\pi_{0.5}$ on LIBERO with 500 evaluation episodes per suite. Synchronous execution is the teacher run at $10{:}0$ with its last 10 actions masked, the input it was trained with (97.60\%, \cref{sec:impl-pi05}). Naive asynchronous execution and DelayOPSD run at $10{:}10$. In every suite DelayOPSD recovers more than half of the gap between naive execution and this teacher, and in Object, Goal and Long it is within 2.20 points of the teacher. Time is the mean duration of successful episodes at the 20\,Hz control rate, plus, for synchronous execution, the measured inference latency of every replanning step (RTX 4090), during which the robot waits. Action jump is the mean Euclidean norm of the change between consecutive executed actions at chunk switches, averaged over all episodes.

\begin{table}[h]
  \centering
  \footnotesize
  \setlength{\tabcolsep}{3pt}
  \caption{\textbf{LIBERO, $\pi_{0.5}$, speed and smoothness} (\cref{fig:teaser}b). Success (\%), time of successful episodes (s) and action jump at chunk switches, per suite and averaged over the four suites.}
  \label{tab:speed}
  \begin{tabular}{@{}llccccc@{}}
    \toprule
    & Method & Spatial & Object & Goal & Long & Avg. \\
    \midrule
    \multirow{3}{*}{Success (\%) $\uparrow$} & Synchronous & 97.40 & 99.80 & 97.80 & 95.40 & 97.60 \\
     & Naive async & 90.80 & 84.00 & 92.40 & 79.00 & 86.55 \\
     & DelayOPSD & 94.40 & 97.60 & 97.00 & 94.40 & 95.85 \\
    \midrule
    \multirow{3}{*}{Time (s) $\downarrow$} & Synchronous & 7.49 & 9.71 & 8.06 & 19.02 & 11.07 \\
     & Naive async & 5.20 & 6.70 & 5.44 & 12.61 & 7.49 \\
     & DelayOPSD & 5.12 & 6.83 & 5.39 & 12.68 & 7.51 \\
    \midrule
    \multirow{3}{*}{Action jump $\downarrow$} & Synchronous & 0.170 & 0.164 & 0.165 & 0.155 & 0.164 \\
     & Naive async & 0.256 & 0.292 & 0.225 & 0.285 & 0.264 \\
     & DelayOPSD & 0.175 & 0.160 & 0.161 & 0.138 & 0.158 \\
    \bottomrule
  \end{tabular}
\end{table}

\subsection{Training for one known delay}
\label{sec:single}

When the deployment delay is known, as on our real robot (\cref{sec:impl-real}), one buffer at that delay suffices. \Cref{tab:single} trains on the $K{=}10$ buffer alone, with the teacher, loss and number of updates of the main experiments. The buffer holds 1,600 delayed rollouts of the base, about as many episodes as the 1,693 demonstrations. No synchronous rollouts are collected and the scene filter is not used, as on the real robot. Failed episodes are cut 30 steps after the stall onset, as in the main experiments (\cref{tab:impl}).

With this one buffer, the student is 5.80 points above naive execution at $K{=}10$ for SmolVLA and 9.55 points for $\pi_{0.5}$, where distilling on the demonstrations with the same teacher gives $-0.25$ and $+3.75$. The student trained on all delays serves every delay with one set of weights, which is the choice when the delay is unknown or differs between deployments.

\begin{table}[h]
  \centering
  \small
  \setlength{\tabcolsep}{4pt}
  \caption{\textbf{Training for one known delay} ($K{=}10$), LIBERO, four-suite average success (\%) at $K{=}10$. Episodes: rollouts or demonstrations used for training.}
  \label{tab:single}
  \begin{tabular}{@{}lccc@{}}
    \toprule
    Training states & Episodes & SmolVLA & $\pi_{0.5}$ \\
    \midrule
    Naive async & -- & 70.40 & 86.55 \\
    Demonstrations & 1{,}693 & 70.15 & 90.30 \\
    One buffer ($K{=}10$) & 1{,}600 & 76.20 & 96.10 \\
    All delays (DelayOPSD) & 17{,}600 & 78.70 & 95.85 \\
    \bottomrule
  \end{tabular}
\end{table}

\subsection{Real robot}
\label{sec:real-more}

\begin{figure*}[tp]
  \centering
  {\small (a) Drawer}\\[2pt]
  \includegraphics[width=\textwidth]{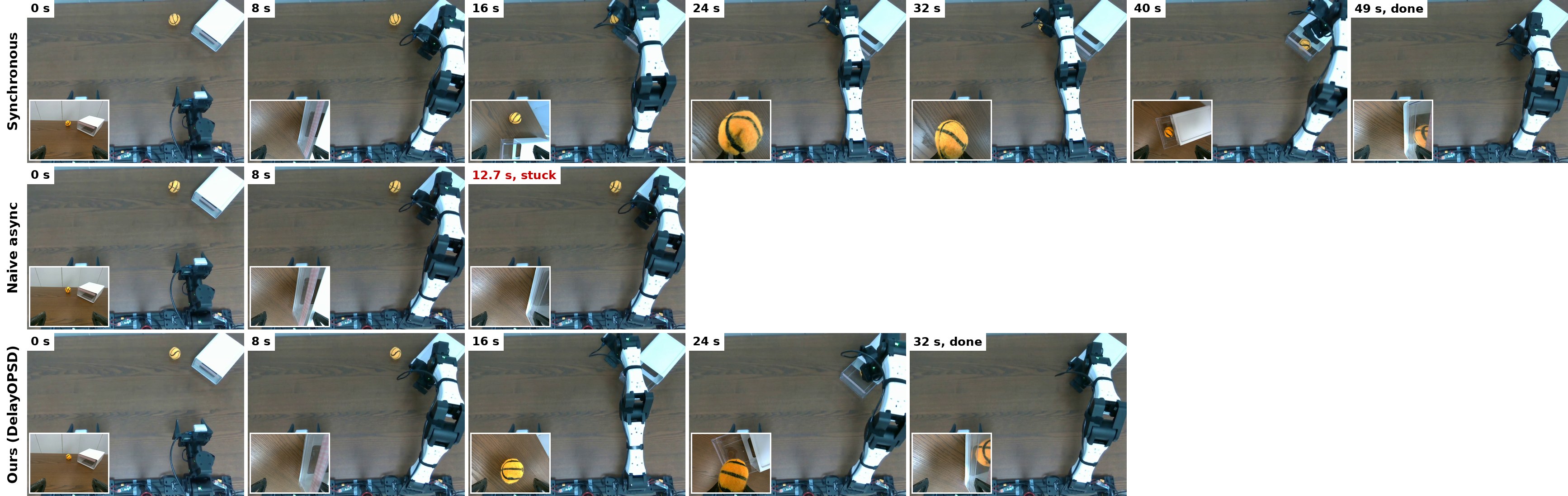}\\[4pt]
  {\small (b) Basket}\\[2pt]
  \includegraphics[width=\textwidth]{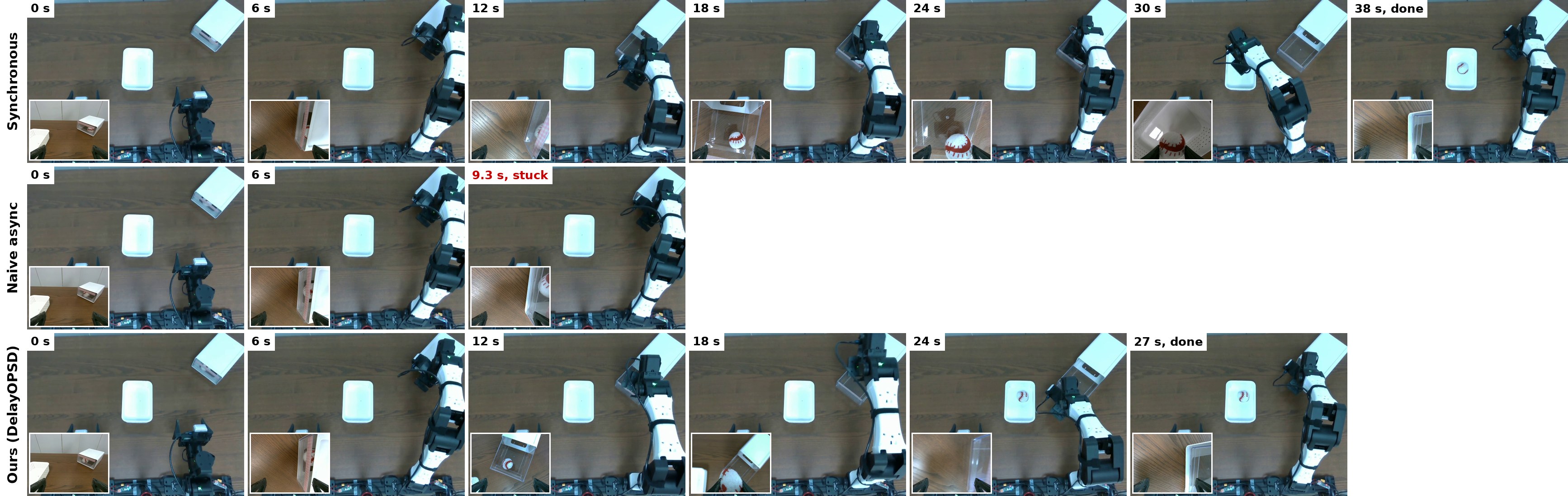}
  \caption{\textbf{Real-robot rollouts from the same scene (scene 4 of each task).} Top-view camera with the wrist camera inset. Synchronous execution succeeds in 49\,s and 38\,s, of which about 14\,s and 10\,s are spent waiting for inference. In both tasks naive asynchronous execution stops against the drawer without opening it. DelayOPSD succeeds in 32\,s and 27\,s. In this scene the three methods succeed in 9, 4 and 9 of 10 trials on Drawer and in 8, 5 and 9 on Basket.}
  \label{fig:real-rollout}
\end{figure*}

The setup and the metrics are defined in \cref{sec:impl-real}, and \cref{tab:real-exec} gives the measured latency. Successful episodes take 30.5/28.4\,s (Drawer/Basket) with
DelayOPSD, 42.1/39.2\,s synchronously, 31.3/29.4\,s naively and 31.1/28.9\,s with RTC. Against naive asynchronous execution, DelayOPSD is higher on both tasks, by 24 points on Drawer and by 8 on Basket, where the delay costs 10 points. It has at least as many successes as naive execution in 9 of the 10 scenes (\cref{tab:real-scenes}). \Cref{fig:real-rollout} shows one scene of each task, and videos of the same trials in real time are at \url{https://github.com/present091/DelayOPSD}.

\begin{table}[h]
  \centering
  \footnotesize
  \setlength{\tabcolsep}{3pt}
  \caption{\textbf{Real-robot execution settings.} End-to-end inference latency per request, measured on the robot side. All methods run on the same GPU. DelayOPSD was run with its LoRA weights not merged into the base, which accounts for its higher latency than naive execution.}
  \label{tab:real-exec}
  \begin{tabular}{@{}llcc@{}}
    \toprule
    & & \multicolumn{2}{c}{Latency (ms)} \\
    \cmidrule(lr){3-4}
    Method & Schedule & median & p99 / max \\
    \midrule
    Synchronous & $15{:}0$, arm held & 192 & 228 / 255 \\
    Naive async & $15{:}15$ & 191 & 225 / 255 \\
    RTC~\cite{rtc} & $15{:}15$ & 412 & 496 / 566 \\
    DelayOPSD & $15{:}15$ & 220 & 256 / 288 \\
    \bottomrule
  \end{tabular}
\end{table}

\begin{table}[h]
  \centering
  \footnotesize
  \setlength{\tabcolsep}{4pt}
  \caption{\textbf{Real robot, successes per scene.} Successful trials out of 10 in each of the 5 scenes of each task. The totals are those of \cref{tab:real}.}
  \label{tab:real-scenes}
  \begin{tabular}{@{}lcccccccccc@{}}
    \toprule
    & \multicolumn{5}{c}{Drawer} & \multicolumn{5}{c}{Basket} \\
    \cmidrule(lr){2-6}\cmidrule(lr){7-11}
    Scene & 1 & 2 & 3 & 4 & 5 & 1 & 2 & 3 & 4 & 5 \\
    \midrule
    Synchronous & 7 & 8 & 3 & 9 & 6 & 6 & 6 & 1 & 8 & 5 \\
    Naive async & 8 & 6 & 3 & 4 & 4 & 5 & 3 & 1 & 5 & 7 \\
    RTC~\cite{rtc} & 6 & 3 & 3 & 8 & 6 & 5 & 5 & 2 & 6 & 5 \\
    DelayOPSD & 8 & 6 & 6 & 9 & 8 & 6 & 4 & 1 & 9 & 5 \\
    \bottomrule
  \end{tabular}
\end{table}

\subsection{Delay measurement on Kinetix}
\label{sec:gap-kinetix}

We repeat the measurement of \cref{tab:gap} on Kinetix with the base policy and the procedure of \cref{sec:gap-details}, over the $R = \max(K, 1)$ executed steps. Expert contexts are drawn from the expert data on which the base was trained. Rollout contexts come from delayed rollouts of the base at delay $K$, and failed episodes keep the tuples left by the failure rule of \cref{sec:impl-kinetix}. Each cell uses 512 contexts per level, and CIs resample contexts within each level.

\begin{table}[h]
  \centering
  \footnotesize
  \setlength{\tabcolsep}{1.5pt}
  \caption{\textbf{Kinetix, change in the executed actions caused by the delay} ($\times 10^{-2}$, mean over 12 levels). Successes and failures are those of the delayed rollouts. Ratios are to the expert data, with 95\% CIs.}
  \label{tab:gap-kinetix}
  \begin{tabular}{@{}lcccc@{}}
    \toprule
    Contexts & $K{=}1$ & 2 & 3 & 4 \\
    \midrule
    Expert data & 1.95 & 2.16 & 2.91 & 3.13 \\
    Successes & 2.84 & 3.75 & 4.62 & 5.87 \\
    Failures & 3.38 & 4.62 & 6.92 & 9.47 \\
    \midrule
    Succ.\ / expert & 1.5 $[1.2, 1.8]$ & 1.7 $[1.5, 2.0]$ & 1.6 $[1.4, 1.8]$ & 1.9 $[1.7, 2.1]$ \\
    Fail.\ / expert & 1.7 $[1.4, 2.1]$ & 2.1 $[1.9, 2.5]$ & 2.4 $[2.2, 2.7]$ & 3.0 $[2.8, 3.3]$ \\
    \bottomrule
  \end{tabular}
\end{table}

As on LIBERO, the delay changes the actions more on the policy's own delayed rollouts than on the data it was trained on, and most in failed episodes (\cref{tab:gap-kinetix}). The ratio grows with the delay and is smaller than on LIBERO (\cref{tab:gap}).

\subsection{Per-suite LIBERO results}
\label{sec:suites-sub}

\Cref{tab:smolvla-suites,tab:pi05-suites} break the four-suite averages of
\cref{tab:smolvla-k,tab:libero-main} down by suite.

\begin{table*}[tp]
  \centering
  \footnotesize
  \setlength{\tabcolsep}{4pt}
  \caption{\textbf{LIBERO, SmolVLA, per suite.} Success rate (\%) per suite (500 episodes each) under $10{:}K$ execution. Avg.\ is the four-suite mean reported in \cref{tab:smolvla-k}. Synchronous execution, the reference: Spatial 85.20, Object 90.20, Goal 88.80, Long 69.20 (avg.\ 83.35). $^\dagger$Privileged state, as in \cref{tab:smolvla-k}.}
  \label{tab:smolvla-suites}
  \begin{tabular*}{\textwidth}{@{\extracolsep{\fill}}llcccccccccc@{}}
    \toprule
    Method & Suite & $K{=}1$ & 2 & 3 & 4 & 5 & 6 & 7 & 8 & 9 & 10 \\
    \midrule
    Naive async & Spatial & 84.60 & 81.60 & 82.80 & 80.40 & 78.00 & 79.00 & 75.60 & 74.80 & 72.00 & 71.60 \\
     & Object & 91.20 & 89.80 & 88.80 & 87.40 & 87.60 & 87.00 & 86.40 & 84.40 & 83.60 & 81.40 \\
     & Goal & 87.60 & 86.60 & 87.40 & 85.00 & 86.80 & 84.00 & 82.60 & 81.60 & 82.00 & 78.80 \\
     & Long & 65.00 & 64.80 & 65.20 & 62.80 & 58.40 & 58.00 & 55.80 & 55.00 & 53.80 & 49.80 \\
     & Avg. & 82.10 & 80.70 & 81.05 & 78.90 & 77.70 & 77.00 & 75.10 & 73.95 & 72.85 & 70.40 \\
    \midrule
    Inf.-time RTC~\cite{rtc} & Spatial & 84.40 & 83.40 & 83.60 & 81.60 & 78.80 & 76.80 & 75.80 & 74.20 & 72.60 & 70.80 \\
     & Object & 90.00 & 90.40 & 91.40 & 87.60 & 87.20 & 86.60 & 85.60 & 85.60 & 83.40 & 82.40 \\
     & Goal & 87.40 & 86.80 & 87.80 & 87.00 & 86.20 & 84.40 & 82.80 & 81.60 & 81.00 & 79.80 \\
     & Long & 65.00 & 67.20 & 66.20 & 62.00 & 57.40 & 60.00 & 58.00 & 54.40 & 54.20 & 50.60 \\
     & Avg. & 81.70 & 81.95 & 82.25 & 79.55 & 77.40 & 76.95 & 75.55 & 73.95 & 72.80 & 70.90 \\
    \midrule
    Train.-time RTC~\cite{trainingtimertc} & Spatial & 85.20 & 83.40 & 82.80 & 82.00 & 80.60 & 81.60 & 83.00 & 80.40 & 79.60 & 81.60 \\
     & Object & 89.20 & 88.80 & 86.00 & 85.20 & 84.20 & 85.60 & 84.80 & 83.40 & 81.80 & 81.60 \\
     & Goal & 85.40 & 83.40 & 84.60 & 84.80 & 84.40 & 85.20 & 84.00 & 83.00 & 84.40 & 83.40 \\
     & Long & 63.80 & 59.40 & 61.40 & 61.20 & 59.80 & 56.80 & 56.20 & 53.60 & 53.80 & 53.60 \\
     & Avg. & 80.90 & 78.75 & 78.70 & 78.30 & 77.25 & 77.30 & 77.00 & 75.10 & 74.90 & 75.05 \\
    \midrule
    VLASH~\cite{vlash}$^\dagger$ & Spatial & 78.20 & 76.00 & 75.80 & 75.80 & 73.20 & 74.80 & 74.60 & 73.40 & 71.40 & 74.40 \\
     & Object & 88.80 & 87.20 & 88.00 & 87.60 & 87.40 & 86.40 & 85.20 & 85.40 & 83.00 & 83.20 \\
     & Goal & 82.60 & 81.60 & 82.40 & 82.60 & 83.20 & 84.00 & 83.80 & 84.40 & 83.80 & 83.40 \\
     & Long & 64.00 & 59.60 & 59.20 & 56.20 & 58.40 & 52.80 & 49.80 & 48.60 & 48.20 & 42.20 \\
     & Avg. & 78.40 & 76.10 & 76.35 & 75.55 & 75.55 & 74.50 & 73.35 & 72.95 & 71.60 & 70.80 \\
    \midrule
    DEFLECT~\cite{deflect} & Spatial & 83.20 & 82.40 & 83.20 & 82.20 & 80.80 & 77.20 & 75.20 & 74.00 & 75.00 & 71.60 \\
     & Object & 89.80 & 90.60 & 89.20 & 87.20 & 88.60 & 85.80 & 86.40 & 82.80 & 82.80 & 80.60 \\
     & Goal & 87.20 & 88.00 & 88.40 & 88.00 & 86.40 & 84.60 & 84.20 & 81.40 & 82.40 & 80.60 \\
     & Long & 66.60 & 65.80 & 64.40 & 60.20 & 60.40 & 58.00 & 57.40 & 52.00 & 52.40 & 49.40 \\
     & Avg. & 81.70 & 81.70 & 81.30 & 79.40 & 79.05 & 76.40 & 75.80 & 72.55 & 73.15 & 70.55 \\
    \midrule
    FutureRTC~\cite{futurertc} & Spatial & 85.60 & 84.80 & 84.40 & 82.20 & 82.60 & 80.00 & 79.80 & 77.40 & 79.40 & 76.20 \\
     & Object & 91.80 & 89.80 & 88.80 & 88.00 & 88.00 & 87.40 & 89.20 & 88.20 & 88.00 & 86.60 \\
     & Goal & 87.60 & 86.20 & 84.00 & 83.60 & 82.20 & 83.00 & 83.00 & 82.40 & 80.40 & 82.40 \\
     & Long & 66.20 & 65.60 & 61.80 & 63.20 & 61.40 & 58.00 & 58.20 & 58.40 & 58.40 & 56.80 \\
     & Avg. & 82.80 & 81.60 & 79.75 & 79.25 & 78.55 & 77.10 & 77.55 & 76.60 & 76.55 & 75.50 \\
    \midrule
    DelayOPSD & Spatial & 84.60 & 84.00 & 84.00 & 83.00 & 83.20 & 80.20 & 80.40 & 80.00 & 78.00 & 78.60 \\
     & Object & 90.40 & 90.40 & 90.80 & 90.80 & 91.80 & 92.20 & 91.60 & 90.40 & 88.20 & 89.40 \\
     & Goal & 87.00 & 85.20 & 85.80 & 87.40 & 84.80 & 83.60 & 84.60 & 84.80 & 82.60 & 83.80 \\
     & Long & 70.60 & 68.60 & 68.60 & 68.20 & 66.40 & 64.20 & 63.00 & 62.40 & 62.00 & 63.00 \\
     & Avg. & 83.15 & 82.05 & 82.30 & 82.35 & 81.55 & 80.05 & 79.90 & 79.40 & 77.70 & 78.70 \\
    \bottomrule
  \end{tabular*}
\end{table*}

\section{Limitations}
\label{sec:limitations}

\paragraph{Rollout collection.} DelayOPSD needs delayed rollouts of the base for every training delay, 17{,}600 episodes on LIBERO against 1{,}693 demonstration episodes (cost in \cref{tab:cost}). They need no human demonstrator, but on a real robot each of them is a physical episode. \Cref{sec:single} trains on the buffer of one delay.

\paragraph{Evaluated delays.} Each evaluation episode uses one delay $K \le R$. Delays that change within an episode, for example under fluctuating network latency, are left for future evaluation.

\paragraph{Teacher errors.} The teacher is the policy itself run synchronously, so the student inherits its errors. In scenes where the synchronous policy also fails, its targets are less accurate (\cref{sec:physical}).

\paragraph{Information the student does not have.} The student does not see the execution-time observation. Changes of the scene during the delay that the committed actions do not reveal cannot be recovered by training (\cref{sec:floor}).

\clearpage
\twocolumn[{%
  \centering
  \footnotesize
  \setlength{\tabcolsep}{4pt}
  \captionof{table}{\textbf{LIBERO, $\pi_{0.5}$, per suite.} Success rate (\%) per suite (500 episodes each) under $10{:}K$ execution. Avg.\ is the four-suite mean reported in \cref{tab:libero-main}. Synchronous base, the reference: Spatial 98.20, Object 98.60, Goal 97.60, Long 96.00 (avg.\ 97.60). $^\dagger$Privileged state, as in \cref{tab:libero-main}.}
  \label{tab:pi05-suites}
  \begin{tabular*}{\textwidth}{@{\extracolsep{\fill}}llcccccccccc@{}}
    \toprule
    Method & Suite & $K{=}1$ & 2 & 3 & 4 & 5 & 6 & 7 & 8 & 9 & 10 \\
    \midrule
    Naive async & Spatial & 98.00 & 97.00 & 96.80 & 96.40 & 93.40 & 93.60 & 93.60 & 92.80 & 91.40 & 90.80 \\
     & Object & 99.20 & 98.00 & 96.80 & 97.00 & 96.60 & 95.80 & 93.60 & 91.80 & 88.80 & 84.00 \\
     & Goal & 97.40 & 97.60 & 96.40 & 97.00 & 96.40 & 93.80 & 93.80 & 93.60 & 93.00 & 92.40 \\
     & Long & 95.60 & 96.20 & 95.40 & 92.40 & 90.80 & 86.60 & 87.20 & 85.40 & 80.40 & 79.00 \\
     & Avg. & 97.55 & 97.20 & 96.35 & 95.70 & 94.30 & 92.45 & 92.05 & 90.90 & 88.40 & 86.55 \\
    \midrule
    Inf.-time RTC~\cite{rtc} & Spatial & 98.00 & 98.20 & 95.40 & 93.60 & 94.00 & 93.60 & 93.80 & 91.40 & 91.00 & 90.20 \\
     & Object & 99.20 & 98.60 & 97.00 & 97.00 & 97.00 & 96.80 & 93.80 & 90.60 & 86.00 & 87.00 \\
     & Goal & 97.80 & 97.60 & 96.00 & 97.80 & 95.60 & 96.00 & 94.40 & 92.80 & 92.40 & 92.40 \\
     & Long & 95.40 & 96.60 & 94.20 & 94.20 & 91.00 & 89.60 & 87.60 & 85.20 & 84.20 & 81.00 \\
     & Avg. & 97.60 & 97.75 & 95.65 & 95.65 & 94.40 & 94.00 & 92.40 & 90.00 & 88.40 & 87.65 \\
    \midrule
    Train.-time RTC~\cite{trainingtimertc} & Spatial & 96.80 & 97.20 & 95.80 & 95.80 & 95.80 & 93.80 & 93.40 & 93.20 & 93.80 & 90.60 \\
     & Object & 99.00 & 99.20 & 98.40 & 98.20 & 98.00 & 97.40 & 97.60 & 95.60 & 93.80 & 92.00 \\
     & Goal & 98.00 & 98.40 & 97.20 & 97.60 & 96.80 & 95.00 & 94.60 & 95.80 & 96.00 & 95.20 \\
     & Long & 94.40 & 96.00 & 95.60 & 95.00 & 92.20 & 91.60 & 90.40 & 88.40 & 86.60 & 84.40 \\
     & Avg. & 97.05 & 97.70 & 96.75 & 96.65 & 95.70 & 94.45 & 94.00 & 93.25 & 92.55 & 90.55 \\
    \midrule
    VLASH~\cite{vlash}$^\dagger$ & Spatial & 90.20 & 92.20 & 90.20 & 88.80 & 88.80 & 86.20 & 85.00 & 83.40 & 81.40 & 78.20 \\
     & Object & 90.40 & 91.40 & 93.60 & 94.20 & 92.00 & 91.40 & 90.40 & 86.80 & 83.80 & 80.40 \\
     & Goal & 87.60 & 88.60 & 88.60 & 87.80 & 87.40 & 89.00 & 86.40 & 83.60 & 82.60 & 78.00 \\
     & Long & 83.60 & 86.60 & 87.60 & 87.80 & 85.20 & 80.60 & 79.80 & 81.40 & 75.60 & 68.80 \\
     & Avg. & 87.95 & 89.70 & 90.00 & 89.65 & 88.35 & 86.80 & 85.40 & 83.80 & 80.85 & 76.35 \\
    \midrule
    DEFLECT~\cite{deflect} & Spatial & 97.60 & 97.40 & 96.60 & 97.00 & 95.00 & 94.00 & 92.40 & 92.60 & 91.00 & 91.00 \\
     & Object & 99.40 & 98.80 & 98.60 & 98.60 & 97.20 & 96.40 & 96.00 & 91.40 & 91.20 & 91.00 \\
     & Goal & 98.60 & 98.20 & 97.60 & 97.40 & 97.80 & 96.60 & 96.00 & 95.20 & 93.40 & 93.60 \\
     & Long & 95.40 & 97.20 & 95.40 & 93.80 & 90.60 & 88.80 & 87.20 & 83.80 & 84.20 & 78.80 \\
     & Avg. & 97.75 & 97.90 & 97.05 & 96.70 & 95.15 & 93.95 & 92.90 & 90.75 & 89.95 & 88.60 \\
    \midrule
    FutureRTC~\cite{futurertc} & Spatial & 97.60 & 97.20 & 97.80 & 97.00 & 96.80 & 97.00 & 96.60 & 95.80 & 97.00 & 96.40 \\
     & Object & 98.60 & 98.80 & 99.20 & 99.20 & 99.00 & 99.40 & 98.60 & 98.20 & 98.60 & 98.20 \\
     & Goal & 98.00 & 98.20 & 97.00 & 97.20 & 95.00 & 95.00 & 95.40 & 95.40 & 93.00 & 93.20 \\
     & Long & 95.60 & 96.40 & 96.00 & 94.80 & 95.80 & 94.60 & 94.60 & 92.80 & 92.20 & 93.20 \\
     & Avg. & 97.45 & 97.65 & 97.50 & 97.05 & 96.65 & 96.50 & 96.30 & 95.55 & 95.20 & 95.25 \\
    \midrule
    DelayOPSD & Spatial & 98.20 & 98.00 & 97.80 & 98.00 & 97.00 & 96.60 & 96.00 & 96.80 & 96.00 & 94.40 \\
     & Object & 99.20 & 99.60 & 99.40 & 100.00 & 98.80 & 99.80 & 99.00 & 99.80 & 98.60 & 97.60 \\
     & Goal & 98.40 & 97.60 & 97.40 & 97.80 & 98.00 & 97.80 & 97.20 & 97.20 & 98.20 & 97.00 \\
     & Long & 94.80 & 94.00 & 93.80 & 95.60 & 94.80 & 95.20 & 94.80 & 95.00 & 93.40 & 94.40 \\
     & Avg. & 97.65 & 97.30 & 97.10 & 97.85 & 97.15 & 97.35 & 96.75 & 97.20 & 96.55 & 95.85 \\
    \bottomrule
  \end{tabular*}
  \vspace{14pt}
}]

\end{document}